%% file: main.tex
\documentclass[10pt,journal,compsoc]{IEEEtran}
\usepackage[utf8]{inputenc}
\usepackage{graphicx}
\usepackage[dvipsnames]{xcolor}
\usepackage{soul,color}
\usepackage{cite}
\usepackage{dirtree}
\usepackage{ragged2e}
\usepackage{subcaption}
\usepackage{float}
\usepackage{tabularx, amsmath}
\usepackage{url}
\usepackage{cleveref}
\usepackage{subcaption}
\usepackage{pgfplots} 
\pgfplotsset{compat=1.17}
\usepgfplotslibrary{groupplots}

\title{Open-Source Framework for Encrypted
Internet and Malicious Traffic Classification}

\author
{
		\IEEEauthorblockN
		{
			Ofek Bader\IEEEauthorrefmark{1}\IEEEauthorrefmark{3},
			Adi Lichy\IEEEauthorrefmark{1}\IEEEauthorrefmark{3},
			Amit Dvir\IEEEauthorrefmark{1}\IEEEauthorrefmark{3},
			Ran Dubin\IEEEauthorrefmark{1}\IEEEauthorrefmark{3},
			Chen Hajaj\IEEEauthorrefmark{2}\IEEEauthorrefmark{3}\IEEEauthorrefmark{4}
			\\
		}
		\IEEEauthorblockA{\IEEEauthorrefmark{1}Department of Computer Science, Ariel University, Israel}\\
		\IEEEauthorblockA{\IEEEauthorrefmark{3}Ariel Cyber Innovation Center, Ariel University, Israel}\\
	    \IEEEauthorblockA{\IEEEauthorrefmark{2}Data Science and Artificial Intelligence Research Center, Ariel University, Israel}\\
	    \IEEEauthorblockA{\IEEEauthorblockA{\IEEEauthorrefmark{4}Department of Industrial Engineering \& Management, Ariel University, Israel}
	    }
	    
	    \IEEEcompsocitemizethanks{\IEEEcompsocthanksitem The authors want to thank Antonio Montieri for his help in the implementation of the original DISTILLER system and also want to thank Tal Shapira for his help in the implementation of the FlowPic feature extraction and its respective model. This work was supported by the Ariel Cyber Innovation Center in conjunction with the Israel National Cyber Directorate in the Prime Minister's Office. }

} 

\begin{document}

\maketitle

\begin{abstract} 
Internet traffic classification plays a key role in network visibility, Quality of Services (QoS), intrusion detection, Quality of Experience (QoE) and traffic-trend analyses. In order to improve privacy, integrity, confidentiality, and protocol obfuscation, the current traffic is based on encryption protocols, e.g., SSL/TLS.
With the increased use of Machine-Learning (ML) and Deep-Learning (DL) models in the literature, comparison between different models and methods has become cumbersome and difficult due to a lack of a standardized framework. In this paper, we propose an open-source framework, named OSF-EIMTC, which can provide the full pipeline of the learning process. From the well-known datasets to extracting new and well-known features, it provides implementations of well-known ML and DL models (from the traffic classification literature) as well as evaluations. Such a framework can facilitate research in traffic classification domains, so that it will be more repeatable, reproducible, easier to execute, and will allow a more accurate comparison of well-known and novel features and models. As part of our framework evaluation, we demonstrate a variety of cases where the framework can be of use, utilizing multiple datasets, models, and feature sets. We show analyses of publicly available datasets and invite the community to participate in our open challenges using the OSF-EIMTC.

\end{abstract}


\section{Introduction}
Internet traffic classification works tackle the internet traffic classification problem from different approaches (e.g., payload-based and behavior-based) and categorize the data representation methods in various ways. For instance, statistical features are taken from the network flows. 
Classical machine-learning models have been shown to be applicable in the scope of internet traffic classification \cite{cite7, cite6, conti_new_2017, cite15}. Several works recently used Natural Language Processing (NLP) techniques such as transforming the flow into a language to use word embedding \cite{cite19}. 

Recently, there has been a huge change in the internet traffic where new network protocols such as QUIC, HTTP/2, HTTP/3 and new privacy concerned protocols such as TLS 1.3 and DoH \cite{rfc8484_doh_dns_over_https, ietf-tls-esni-13, ietf-quic-http-34, importance_esni_circumvention, ietf-tls-rfc8446bis-03} have been introduced. Consequently, the ability of the well-known solutions, which are based on DPI, ML, and DL classification systems, will be affected and will require extensive research. 

Regarding any network classification problem based on the ML/DL pipeline, the researchers must cope with several important questions: beginning with which dataset should be used, how to extract a set of features, how to construct a new model, and how to compare the new model with the well-known models. Due to the lack of a common shared framework for internet traffic classification, the above tasks are difficult and tiresome. For example, in the application network traffic classification task \cite{FlowPic2021, Wang1DCNN, Distiller, LopezMartin, BOA_conf}, researchers would like to compare their features and models using the same framework, especially in systems using the same subset of features (e.g., 784 payload bytes of the flow) \cite{Distiller, Wang1DCNN, maldist_ccnc, DeepMAL} and minor changes in the models such as a similar CNN architecture \cite{FlowPic2021, WeiWangMalwareTrafficClassification}.

\begin{figure*}[!]
    \centering
    \includegraphics[width=1\textwidth, height = 1.75 cm]{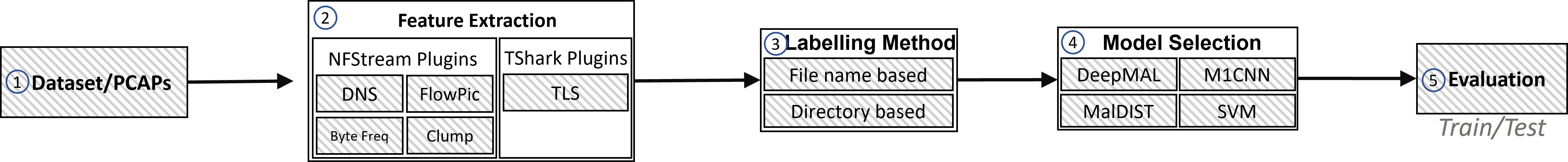}
    \caption{The Open-Source Framework for Encrypted
Internet and Malicious Traffic Classification Pipeline}
    \label{fig:framework-complete-flow}
\end{figure*}

Therefore, the contribution of this work is as follows: we introduce a novel Open-Source Framework for Encrypted Internet and Malicious Traffic Classification (OSF-EIMTC). The OSF-EIMTC provides a full ML/DL pipeline as can be depicted in Figure \ref{fig:framework-complete-flow}. The pipeline can be characterized by the following phases: dataset selection, feature extraction, model selection, and evaluation. In the dataset selection phase, the framework allows access to well-known datasets (e.g., ISCXVPN2016, Ariel). In the feature extraction phase, the framework can extract sets of state-of-the-art features (e.g., flow and packet payload bytes \cite{WeiWangMalwareTrafficClassification, Wang1DCNN, Distiller, maldist_ccnc, DeepMAL}), with the ability to add new feature sets (via plugins). In the model selection phase, the framework also allows access to standard ML classifiers (e.g., RF, SVM) and provides implementations of state-of-the-art deep-learning classification models (e.g., MalDIST \cite{maldist_ccnc}). In the evaluation phase, the framework implements standard model evaluation metrics (e.g., accuracy, recall). The framework can be for researchers as a benchmarking platform to evaluate their approaches by comprehensive to well known features and solutions. 

Furthermore, researchers can extend the framework platform with new features and models in order to contribute to the research community. We utilize the framework to evaluate different models over different datasets, showcasing the framework's advantages. We also used the framework to analyze several public state-of-the-art datasets and providing several interesting insights about the datasets. 

The rest of this paper is organized as follows. Section \ref{sec:related-work} describes the related works. Section \ref{sec:framework} discusses the framework implementation. Section \ref{sec:dataset-analysis} presents various datasets and their analysis. Section \ref{sec:dry-run-evaluation} presents runs of the framework evaluation. Section \ref{sec:online-challenge-evalai} presents online challenges, and in conclusion, section \ref{sec:summary} presents a summary and discusses future work.

\section{Related Work} \label{sec:related-work}
In recent years, deep-learning models have become the prominent method for network traffic classification. Some works that utilize deep-learning such as \cite{WeiWangMalwareTrafficClassification, Wang1DCNN, citeMTATLS} even utilize raw data of the payload of the flow to feed deep-learning algorithms, thereby demonstrating that hand-crafting features are not always needed. 

The works that utilize deep-learning span over multiple scopes and domains, with works that focus on the classification of the operating system, browser, and application levels \cite{cite7}; mobile app identification \cite{cite6, conti_new_2017}; and even webpage fingerprinting \cite{cite15}.  Several works converted the network flow into an image to harness image processing techniques and equivalent Deep-Learning (DL) architectures \cite{ cite3, cite6, cite10, FlowPic2021, WeiWangMalwareTrafficClassification, DeepMAL}. Newer works incorporated the Ordinary Differential Equation Network (ODENet) within a DL architecture to classify uni-directional network flows \cite{ODENETFastLeanEncryptedClassification}.
If we shift the focus to the cyber domain, many works tackle the task of malware network traffic detection and classification. \cite{WeiWangMalwareTrafficClassification, NetML, UnknownMalwareDetectionBenGurion, EncryptedMalwareContext, MTAKDD19, DeepMAL, yesML, cite27, cite29, cite30, citeMTATLS}.

Along with detection or classification methods, the data in use have great importance as well. Some publicly available datasets such as ISCXVPN2016 are widely used in the literature in works such as \cite{Wang1DCNN, Distiller, FlowPic2021, NetML, ISCX2016, maldist_ccnc, ODENETFastLeanEncryptedClassification}. Where the works adopt the dataset for the purpose of classifying a network flow according to its encapsulation type (tunneled via VPN or not), traffic type (e.g., video/audio/chat), and application (e.g., Skype/Netflix). However, the dataset contains extensive background noise \cite{MT_21}, such as unrelated BlueStacks \cite{Distiller} and Dropbox \cite{maldist_ccnc} network traffic. Researchers may need to clean and preprocess the data before using it. Therefore, access to several clean datasets in one place is an advantage for researchers. For example, Barut el. al., \cite{MT_21} 
compiled their dataset with well-known datasets for both malware detection and application type categorization.

As many works attempt to solve similar tasks, researchers may want to compare their own models and features to other proposed traffic classification methods as part of the evaluation process. NetML \cite{NetML} is an example of a framework that provides its own compiled data files from publicly available sources with features already extracted and ready to use. In both normal and malicious network traffic tasks, researchers need to implement their own models and compare them to others while using the same provided feature files. The features were extracted with an accelerated feature extraction library by Intel. However, access to the library and the code used for feature extraction purposes are not given or shared. This complicates the ability to replicate the same features and methods with another dataset. Moreover, NetML does not implement any state-of-the-art deep-learning (DL) models. nPrint \cite{nPrintAutomation} is another tool that unifies the representation of each packet into a standard presentation that is amenable for representation learning. Though nPrint is designated to automate the process of machine learning pipelines, it does not propose an option for the extraction of custom features. While nPrint is integrated with an AutoML library (AutoGluon-Tabular), it does not support user-defined models, such as Deep-Learning neural networks (e.g., M1CNN \cite{Wang1DCNN}, MalDIST \cite{maldist_ccnc}).

\section{The Framework} \label{sec:framework}
Our framework is an open-source \cite{OSFEIMTC_GitHub} that enables a comparison of multiple new and well-known features and state-of-the-art models for both network and malware traffic classification. The proposed framework allows the researcher to acquire a complete ML/DL pipeline with minimal time and effort as a benchmark to compare their new ideas and as a platform to plugin new features and models.

The envisioned flow of the ML/DL pipeline is presented in  Figure~\ref{fig:framework-complete-flow}, demonstrating the ability of the framework to create a complete pipeline scheme The workflow comprises a total of 5 parts:

\begin{enumerate}
    \item Data: The network traffic data files match the task that the researcher is attempting to solve, whether it be VPN detection, application network classification, or malicious traffic identification. The data is one of the most crucial components in research and experiments to achieve satisfactory results in real-world scenarios. \textbf{Our framework provides access to some of the well-known datasets e.g., \cite{ISCX2016_Dataset, MTA, stratodatasets, USTC} in an organized manner.} Some details about the well-known data sources can be found in Section \ref{sec:framework-datasets} while their analysis can be found in Section \ref{sec:dataset-analysis} and in the Appendix.
    
    \item Feature extraction: Many of the previous works used different tools for feature extraction, where some of them are not publicly available \cite{NetML}. The lack of public access to the feature extraction tools affects the reproducibility of the same features. Furthermore, the existence of different tools complicates the process of extracting features, and makes it difficult to combine different feature sets from different tools. Moreover, currently combining new features with old ones is a complicated task. \textbf{The framework can extract various well-known features, for example, flow-based and packet size-related features (e.g., min, max, and mean of packet sizes of the flow). The framework can also extract TLS-related features such as TLS record size and direction as in \cite{encrypted_http2_time_space_2020, Decetion_ENC_MAL_ML, DoHTunnels}. Additionally, the framework is able to extract the well-known full features set of well-known works such as DeepMAL \cite{DeepMAL} payload bytes per packet, FlowPic images \cite{FlowPic2021}, and the features for the modalities of DISTILLER \cite{Distiller} and MalDIST \cite{maldist_ccnc}. Given the ability to extract standalone features (e.g., TLS features) and full feature sets such as FlowPic, facilitates the ability to create new feature sets that are a combination of well-known standalone features, and to add new features.} More details on feature extraction are provided in Section ~\ref{sec:feature_extraction}.
    
    \item Labelling method: every sample should have a respective label. Every feature set extracted from the sample should have the same label. The labelling part takes the extracted features for each sample and merges them with their respective label. By providing a rule or a naming scheme, it is possible to automatically associate any feature set created from flows/packets (or any other network sample) with their labels. The respective labelling can be done in several ways such as correlating them with their original files or directories. 
    \textbf{The framework allows one to configure the labeling method with the suitable naming scheme}. See Section ~\ref{sec:labeling_methods} for examples.
    \item Model selection: Many models exist such as classical machine-learning and deep-learning neural networks. A comparison should be conducted for any new model (which can be based on older models). Therefore, \textbf{the framework provides classical ML/DL models and implementations of state-of-the-art deep-learning architectures such as MalDIST \cite{maldist_ccnc} for malware detection \& classification and M1CNN \cite{Wang1DCNN} for application classification}. The models are discussed in detail in Section ~\ref{sec:models}.
    
    \item Evaluation: To determine the performance and predict the robustness and effectiveness of a model, along with its provided features, on a particular task requires accurate evaluation methods. Comparison of a variety of models is crucial for quality research work. \textbf{In order to make the process of a new evaluation or model comparison easier, common evaluation metrics (such as accuracy, recall, precision, and F1-score) are provided in the framework.}
\end{enumerate}

In the upcoming sections, we elaborate on the core parts of the framework pipeline, which include datasets, feature extraction, labeling methods, and model selection.

\begin{table}[h!]
\centering
 \begin{tabularx}{1\linewidth}{|p{0.17\linewidth}|p{0.13\linewidth}|p{0.45\linewidth}|p{0.05\linewidth}|}
 \hline
 Dataset & Domain & Contents & Ref. \\ 
 \hline 
 USTC2016 & Malware, Apps & 10 types of malware families and 10 consumer applications. & \cite{WeiWangMalwareTrafficClassification, DeepMAL, maldist_ccnc} \\\hline
 Stratosphere & Malware &  Variety of captured and/or simulated malware traffic, some captures are mixed (benign and malware) and the rest are purely benign. & \cite{MT_21, NetML, maldist_ccnc} \\\hline
 MTA & Malware & Variety of malware network samples. & \cite{MTAKDD19, maldist_ccnc} \\\hline
 ISCX2016 & Apps & Samples of applications of different categories, along with VPN encapsulated traffic. & \cite{NetML, ISCX2016, Distiller, Wang1DCNN, maldist_ccnc, MT_21, ODENETFastLeanEncryptedClassification}, \\\hline
 Ariel (BOA2016) & Browsers, OS, Apps & Samples of web applications traffics labeled by browser and operating system. & \cite{BOA_conf, QUICClassificationFewLabels} \\\hline
 MAppGraph 2021 & Mobile Apps & Samples of over 80 popular mobile applications in Google Play. & \cite{MAppGraph2021} \\\hline

 \hline
 \end{tabularx}
\end{table}

\subsection{Datasets} \label{sec:framework-datasets}
The framework provides a catalog of datasets for easy access. In this section, we describe the contents of selected datasets.
We analyze some of the datasets in section \ref{sec:dataset-analysis}.

\subsubsection{USTC2016 \cite{USTC}}
This dataset, which can be found here in \cite{USTC}, contains 10 malware families and 10 types of benign traffic.

\begin{itemize}
    \item The malware classes are:  Cridex (Dridex), Geodo (Emotet), Htbot, Miuref, Neris, Nsis-a, Shifu, Tinba, Virut, Zeus.
    \item The benign classes are: BitTorrent, Facetime, FTP, Gmail, MySQL, Outlook, Skype, SMB, Weibo, WorldOfWarcraft.
\end{itemize}

Some works that used this dataset can be found in \cite{maldist_ccnc, WeiWangMalwareTrafficClassification, DeepMAL}

\subsubsection{StratosphereIPS\cite{stratodatasets}}
  Is a dataset comprising of three parts: benign, malware, and mixed traffic. The dataset was generated by Stratosphere Laboratory, a part of the CTU University of Prague in the Czech Republic. Additional information such as a description of the behavior captured facilitates the labeling process. Some works that adopt this dataset can be found in  ~\cite{Ransomware_pre-encrypted_alert,Decetion_ENC_MAL_ML,NetML}
 
\subsubsection{MTA (Malware Traffic Analysis) \cite{MTA}}
 The data source is a website (blog) that includes many types of malware infection traffic for analysis. The website contains many types of malware such as ransomware and exploit kits. As of 2013 to date, the blog is updated daily with relevant malware traffic, continuously adding more samples to the dataset. Using Intrusion-Detection Systems (IDS) and Antivirus software, every binary file in the PCAPs has been confirmed as malicious. Papers such as \cite{citeMTATLS, MTAKDD19, maldist_ccnc} used this dataset for malware detection.
 
\subsubsection{ISCX2016 (ISCXVPN2016) \cite{ISCX2016_Dataset}}
This dataset consists of 150 PCAP files of different types of traffic and applications. Each PCAP file has an application category (e.g., Spotify, Facebook, YouTube, etc.), a traffic type category (e.g., streaming, VoIP, chat, etc.) and an encapsulation label (non-VPN/VPN). Some works that adopted this dataset can be found in \cite{FlowPic2021,NetML,Distiller, maldist_ccnc, Wang1DCNN, MT_21, ODENETFastLeanEncryptedClassification}.

Note that the data contains extensive noise, such as unrelated BlueStacks traffic, Dropbox traffic, and background traffic. Works such as \cite{FlowPic2021} extracted only the relevant flows from each PCAP file, whether the audio flow of Skype or the video flow of Netflix. Many works dropped browser traffic files due to lack of sufficient accurate labeling, and p2p traffic such as BitTorrent due to lack of counterparts in non-encapsulated traffic.

\subsubsection{Ariel (BOA2016) \cite{BOA_conf}}
This dataset is from a paper in which the authors collected the data over a period of more than two months, in their lab, using a selenium web crawler for browser traffic. The dataset contains applications' traffic such as YouTube, and Facebook, which are labeled as browser traffic, and Dropbox and TeamViewer that are labeled as non-browser traffic. The dataset contains more than 20,000 sessions. The average duration of a session was 518 seconds where on average each session had 520 forward packets (the average forward traffic size was 261K bytes) and 637 backward packets (the average backward traffic size was 615K bytes). Almost all of the flows are TLS encrypted. Examples of works that used this dataset include \cite{BOA_conf, QUICClassificationFewLabels}.

\subsubsection{MAppGraph dataset (2021) \cite{MAppGraph2021}}
The authors collected traffic of 101 \textbf{mobile applications}, which are popular in Google Play. For each application, more than 30 hours of traffic were collected, resulting in more than 600 GB of traffic stored in PCAP files. However, the authors published a portion of the dataset, which amounts to 81 applications that weighs around 500 GB. Some of the applications include but not limited to: Facebook, Twitch, Instagram, Skype, Spotify, Google Meet, Soundcloud, and Zoom.

\subsection{Feature extraction} \label{sec:feature_extraction}
Features are one of the key components in the success of a machine learning model. There are many scopes of features; one can extract features from a packet (e.g., size and time), protocol/header (e.g., type and information), uni-directional flow (e.g., number of uploaded or downloaded bytes), bi-directional flow (e.g., average packet size), or a time window. The feature extraction of the framework is used to extract features from packets/flow/session. Hence, we can also extract features from uni and bi-directional flows. In this section, we present some standalone features and features sets that the framework supports. Note that, the framework offers the flexibility of adding any arbitrary feature or feature set that can be defined and harvested from the sample.

Our feature extraction component is based on the NFStream package~\cite{NFStream}. NFStream is a Python framework that provides a fast, flexible, and expressive data structures designed to make working with PCAP files both easy and intuitive. Some of the features of nDPI, which is an open and extensible deep packet inspection library, are integrated into NFStream and provides additional flow information such as fingerprints, request server name, and application category detection. To further enrich the of types of features extracted, we utilize TShark~\cite{tshark} as a post-NFStream tool to extract TLS related features per flow.

We developed new plugins to extract state-of-the-art features and their preprocessing phase, to make it easier to compare different models and extensions to allow new research directions with ease.

As the feature selection is provided in 'plugins', one can select a set of plugins with various features sets to extract and assemble the required final feature data file. The users can extend the component by developing their own plugins. We envision that more researchers would publish their code using these plugins to make the experiments more easily reproducible and therefore facilitate their use in different scenarios with far greater ease. 

\begin{table}[h!]
\centering
 \begin{tabularx}{1\linewidth}{|p{0.22\linewidth}|p{0.20\linewidth}|p{0.33\linewidth}|p{0.05\linewidth}|}
 \hline
 Plugin & Description & Features & {Ref.} \\ 
 \hline 
 ASN\_info & Autonomous System & ASN number, country code and description & \cite{RobustDetectionNitay, UnknownMalwareDetectionBenGurion} \\\hline
 DNS & Domain Name System & \#IP addr in response, TTL, domain name text statistics. & \cite{EncryptedMalwareContext, UnknownMalwareDetectionBenGurion, DeNAT} \\\hline
 n\_bytes & Payload bytes of the first N bytes of the flow. & array of length of N with the value of each byte. & \cite{Wang1DCNN, WeiWangMalwareTrafficClassification, maldist_ccnc, Distiller} \\\hline
 ByteFrequency & Byte distribution of the first N packets. & array of length of 256 with the distribution of each byte. & \cite{EncryptedMalwareContext, TheChallengeClumpSubFlows} \\\hline
 SmallPacketRatio & The ratio of small packets from all the packets & Decimal ratio (single number) & \cite{BotnetSmallPacket, MTAKDD19} \\\hline
 Protocol Headers & IAT, size, direction TCP-win size of packets & A $(n,4)$ matrix, 4 features per packet of $n$ packets. & \cite{LopezMartin, Distiller, maldist_ccnc} \\ \hline
 Clumps & Statistical features over consecutive packets in the same direction. & Min, max, mean, stddev of clump lengths, sizes, and IAT. & \cite{DoHTunnels, TheChallengeClumpSubFlows}\\ 
 \hline
 \end{tabularx}
 
 \caption{A subset of plugins provided in the framework used for feature extraction.}
\end{table}

\textbf{State-of-the-art features plugins:}
While the framework has numerous different plugins, we only discuss a subset of the plugins developed as part of the feature extraction component.
\begin{itemize}
\item ASN\_info (Autonomous system information)~\cite{RobustDetectionNitay, UnknownMalwareDetectionBenGurion}: This plugin extracts and attaches ASN related information such as ASN number, ASN country code and ASN description to each flow. For example, the following ASN information will be attached to a biflow with the source IP address 131.202.240.87 and the destination IP address 178.237.19.228:
    \[
    [num, code, description]
    \]
    \textbf{[num=611, code=CA, description=NB-PEI-EDUCATION-COMPUTER-NETWORK - University of Toronto]} for the source IP and 
    \textbf{[num=47764, code=RU, description=MAILRU-AS Mail.Ru]} for the destination.
    
    Note that the information is pulled from a local file-based DB, which may need to be downloaded and updated from time to time.
     
    \item DNS \cite{EncryptedMalwareContext, UnknownMalwareDetectionBenGurion, DeNAT}:
    Malicious actors have utilized Command \& Control (C2) communication channels over the Domain Name Service (DNS) and, in some cases, have even used the protocol to exfiltrate data. Malicious actors have also infiltrated malicious data/payloads to the victim's system over the DNS.
    
    Malware uses DNS as a way to bypass the block of the C2 server's IP address when it becomes known that the IP address is compromised or is used for malicious intents. The DNS along with a domain name for the C2 server allows the malware to communicate with its C2 server when the malware changes its IP address. 
   Some works that tackled the detection of encrypted malware traffic, such as \cite{EncryptedMalwareContext}, utilized features from DNS traffic such as the number of IP addresses returned by a DNS response, the amount of digits in a DNS response, TTL values, and the domain name length.
   The plugin extracts a wide variety of DNS-related features, such as the number of answers per type, Time-To-Live values, and character-based counters in the answer section, for example digit count, hyphen count, dot count, and more.
   
   Other works such as \cite{DeNAT} used DNS parameters to try to tackle the classification of users behind NAT, as part of internet traffic classification.
    
    \item Early raw byte payload (n\_bytes) \cite{Wang1DCNN, WeiWangMalwareTrafficClassification, Distiller, maldist_ccnc}: This plugin extracts the first n bytes of the flow's payload, which can span over multiple packets in both upload and download links. Works such as \cite{WeiWangMalwareTrafficClassification, maldist_ccnc} used the raw data of the flow to detect and classify malware families by their network traffic, while \cite{Wang1DCNN, Distiller} used it for encrypted network traffic classification tasks.
    \item Byte Frequency \cite{EncryptedMalwareContext, TheChallengeClumpSubFlows}: This plugin extracts the frequency of each byte value ([0-255]) that appears in the payload of the first N packets.
    \item Small Packet Ratio \cite{BotnetSmallPacket, MTAKDD19}: A statistical feature of 
    \[
    \frac{\#small\_packets}{\#packets} 
    \]
    where $\#small\_packets$ is the number of packets with small payload ($\leq threshold$).
    The small packet ratio was used by Hung and Sun~\cite{BotnetSmallPacket} for the detection system of botnets and it was presented in \cite{MTAKDD19}.
    \item Protocol Headers \cite{LopezMartin, Distiller, maldist_ccnc}: As proposed for example in \cite{LopezMartin}, this plugin extracts size, IAT (Inter-arrival time), direction and TCP win-size from the first $n$ packets of the flow to form a $(n,4)$ sized matrix. Lopez-Martin et al. proposed to use $n=20$ while \cite{Distiller} used $n=32$. A lower number of packets may be used for time critical tasks such as detecting malicious traffic. Lopez-Martin et al. \cite{LopezMartin} analyzed the performance of a specific deep-learning model with different number of packets given during training, for the task of differentiating IoT services, and showed that there is no significant impact on the model's performance even with as little as 6 packets. 
    
    In another paper, Rezaei et al.\cite{cite14} used the 6 first packets of the flow to identify mobile applications, the first packets in a TLS flow are plain and contain meaningful information.
    \item SubFlows/Clumps \cite{DoHTunnels, TheChallengeClumpSubFlows}: This plugin extracts clump/subflow related features for each direction. It aggregates the packets of a session into clumps (or subflows), where a clump/subflow is a group of one or more consecutive packets with the same direction. It then extracts data about those clumps such as the min, max, and mean of clump's sizes (number of packets in a clump) and clump's lengths (number of bytes), and internal clump data such as min, max, mean of packets lengths, along with inter-arrival time and more. The rationale behind such grouping is that the application traffic is scattered among several packets as part of the process of TCP segmentation. Mohammadreza et al. \cite{DoHTunnels} used the clumping method in order to lower the dimensionality of data. In contrast, Hong-Yen et al. \cite{TheChallengeClumpSubFlows} named such clumps subflows and claimed that their characteristics can enable machine-learning models to more accurately distinguish different types of sections in a flow while trying to detect changing points of switching applications over VPN encrypted traffic.

\item TLS features \cite{DoHTunnels, IncrementalLearning}: As the use of encryption in network traffic has become popular in recent years, TLS based features can be beneficial for the classification of encrypted flows. The TLS protocol is built on top of a reliable transport protocol such as TCP or QUIC. The data is transferred in the form of TLS records, where one of the first records in the flow is the "Client Hello" type from the client to the server, to which then the server responds with a "Server Hello" record of its own; these records are not encrypted (without eSNI, ECH and up until TLS 1.3) and contain a vast configuration data of the flow, such as cipher suites, the application layer protocol and many more. Some works such as \cite{DoHTunnels} and \cite{IncrementalLearning} used the type and number of cipher suits offered by the client in order to detect TLS traffic initiated by malware, as different TLS clients use or support a different default or predefined set of cipher suits. The information inside the non-encrypted records such as the list of available compression methods, the list of available TLS extensions, and the list of available cipher suits, can be used to generate a client fingerprint. One of the methods for such fingerprinting is JA3 \cite{JA3}. Other works such as \cite{BOA_conf} used the SNI field to classify and label network flows of services/applications such as Twitter and Youtube, where \cite{NetML} extracted 14 different TLS features to classify network attacks, applications, and traffic types.
\textbf{Extraction of TLS features such as TLS record sizes, types and TLS clumps are embedded in the framework by utilizing TShark as a step inside the feature extraction process. }

\end{itemize}

\subsection{Implementation of Common Labeling Methods} \label{sec:labeling_methods}
The proposed framework allows the use of two well-known name based labeling methods, a file name based and a directory based labeling inspired by the TensorFlow's \cite{tensorflowDevelopers2021} built-in utility function of loading an image dataset from a directory \cite{tensorflow_load_image_dataset_from_dir}.
\begin{enumerate}
    \item \textbf{File name based labeling method:} public datasets such as ISCX2016 use filename based labeling, such as a PCAP file, where the file name "vpn\_facebook\_audio2.pcap" has the following labels: \textit{encapsulation type = vpn, traffic type = audio, application = facebook}. An example of a file/directory structure is as follows:
    
    \begin{flushleft}
    \dirtree{%
        .1 data.
        .2 vpn\_facebook\_audio2.pcap.
        .2 skype\_video2b.pcap.
        .2 vpn\_bittorrent.pcap.
    }
    \end{flushleft}

    \item \textbf{Directory based labeling:} inspired by the TensorFlow's utility function for loading an image dataset (tf.keras.utils.image\_dataset\_from\_directory)\cite{tensorflow_load_image_dataset_from_dir}. PCAP files can be arranged into directories, where the directory name determines the label of all files in it. This is useful when the data for each label is too large to be contained in a single file or spans across many different files. An example of a file/directory structure is:
    
    \begin{flushleft}
    \dirtree{%
        .1 data.
        .2 audio.
        .3 vpn\_facebook\_audio2.pcap.
        .3 voipbuster\_4b.pcap.
        .2 video.
        .3 skype\_video2b.pcap.
        .2 p2p.
        .3 vpn\_bittorrent.pcap.
    }
    \end{flushleft}
\end{enumerate}

\subsection{State-Of-The-Art Models} \label{sec:models}
While the users can utilize the extracted features data for training with any classical machine learning model available (e.g., K-Nearest-Neighbors (KNN), Support-Vector-Machine (SVM) and Random Forest (RF)), we also provide state-of-the-art deep-learning models \cite{DeepMAL,WeiWangMalwareTrafficClassification,Wang1DCNN, FlowPic2021, Distiller,maldist_ccnc }.

\begin{itemize}
    \item DeepMAL \cite{DeepMAL}: Marin et al. proposed two variants for malware network traffic detection, one is packet-based and the other is flow-based. The packet-based variant trains and predicts the data extracted from single packets, while the flow-based variant requires multiple packets from the same flow. The DeepMAL raw flows model (flow-based variant) is trained on an input of $n$ \textbf{payload} bytes per packet for the first $m$ packets of the session. the input size is $(m,n)$ per instance. the plugin extracts $(m,n)$ features per session as a matrix, where each byte is represented as a decimal value in $[0,255]$.
    \item M2CNN \cite{WeiWangMalwareTrafficClassification}: Wang et al. proposed a custom LeNet-5 neural-network architecture for malware network traffic detection and classification by leveraging the ability of 2D convolution layers to recognize different image-like patterns by feeding the model named by us as M2CNN, with a matrix-shaped input of bytes (an 8-bit gray image). M2CNN is trained on the $n$ first \textbf{payload} bytes of the flow. M2CNN takes the bytes as a matrix of size $(\sqrt{n}, \sqrt{n})$. The n\_bytes plugin extracts the first $n$ \textbf{payload} bytes of the session ($n$ features), where each byte is represented as a decimal value in $[0,255]$.
    \item M1CNN \cite{Wang1DCNN}: Wang et al. proposed viewing the bytes not as an image, but as a sequential time series, so the authors represented the bytes as a single array of length of $784$ and utilized 1D convolutional layers instead of 2D, aiming to tackle the task of normal network traffic classification. M1CNN is trained on the $n$ first \textbf{payload} bytes of the flow, the M1CNN takes the bytes as a single array of size $(1,n)$. the n\_bytes plugin extracts the first $n$ \textbf{payload} bytes of the session ($n$ features), where each byte is represented as a decimal value in $[0,255]$.
    \item FlowPic \cite{FlowPic2021}: Shapira and Shavitt used a custom LeNet-5 neural network architecture. A FlowPic is a 2D-histogram-image of \textbf{IP} packet sizes and relative time-of-arrival. The plugin creates a FlowPic for each defined time window in the session and saves it on the file system as a compressed NumPy file (.npz), if the session has multiple time windows, multiple FlowPics can be created.
    \item DISTILLER \cite{Distiller}: Aceto et al. proposed a multi-task multi-modal deep-learning model used for classification of network traffic flows. the input to the model is of size $912$, where the first $784$ features are the $784$ payload bytes of the flow, and the next $128$ features are the features proposed by Lopez-Martin et al. \cite{LopezMartin} of [Size, IAT, direction, TCP win-size] of the first $32$ packets.
    \item MalDIST \cite{maldist_ccnc}: In \cite{maldist_ccnc}, the authors proposed a DISTILLER based variant that is used for malware detection and classification. The variant has a third novel modal of shape $(5,14)$ inspired by the STNN model \cite{STNN}, resulting in a total number of $982$ features. In the third modal, the packets in each flow are categorized into 5 categories, where then 14 statistical features are extracted from each category. The features of each category are then aligned in a row, resulting in an image with a shape of (5,14).
    \item A Custom DISTILLER Variant: One may wish to build a customized DISTILLER variant with new and/or different modalities. Since the model is a multitask model, i.e., predicts multiple classifications for the same flow, it is also possible to configure the number of tasks that the model is aimed for.
\end{itemize}

For each of the state-of-the-art models, we can use the model as standalone and the framework also provides the plugins that extracts their required features from raw PCAP files along with \textbf{the necessary preprocessing steps} that are required to perform before feeding it into the model for training and predicting.

\section{Framework Datasets Analysis} \label{sec:dataset-analysis}
In this section, we focus on some state-of-the-art publicly available datasets. We analyze each dataset with some insightful characteristics and bring conclusions. The datasets vary in their domains, from malware traffic to VPN-tunneled traffic. 

We decided to present here one of the dataset analysis (USTC2016), while other three can be found in the Appendix. The main reason for using the framework to analyze the datasets is to shed light on the disadvantages of the well-known datasets, allowing the scientific community take these insights into account when using them. Moreover, we encourage the community to publish new updated, and diverse datasets.

\begin{figure*}[h!]
    \begin{minipage}[t]{0.98\linewidth}
        \centering
        \includegraphics[scale=0.5]{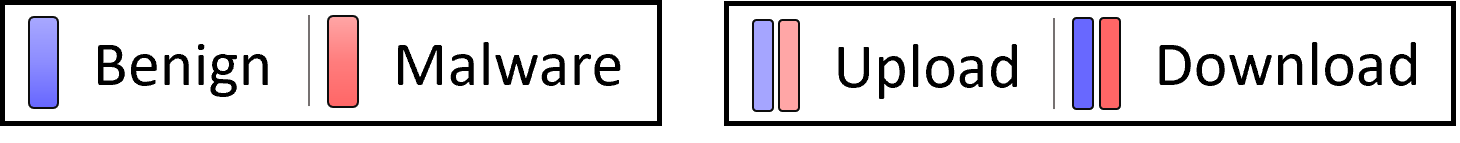}
    \end{minipage}
    
    \begin{minipage}[t]{0.98\linewidth}
        \centering
        \input{assets/tikz/ustc/ustc_avg_packet_size.tikz}
        \phantomsubcaption
        \label{app:fig:ustc_avg_pkt_size}
    \end{minipage}
    \begin{minipage}[t]{0.98\linewidth}
        \centering
        \input{assets/tikz/ustc/ustc_avg_packets.tikz}
        \phantomsubcaption
        \label{app:fig:ustc_avg_pkts}
    \end{minipage}
    \begin{minipage}[t]{0.98\linewidth}
        \centering
        \input{assets/tikz/ustc/ustc_avg_bytes.tikz}
        \phantomsubcaption
        \label{app:fig:ustc_avg_bytes}
    \end{minipage}
    \begin{minipage}[t]{0.98\linewidth}
        \centering
        \input{assets/tikz/ustc/ustc_avg_duration_s.tikz}
        \phantomsubcaption
        \label{app:fig:ustc_avg_dur_s}
    \end{minipage}
    \begin{minipage}[t]{0.98\linewidth}
        \centering
        \input{assets/tikz/ustc/ustc_avg_iat_ms.tikz}
        \phantomsubcaption
        \label{app:fig:ustc_avg_iat_ms}
    \end{minipage}
    \begin{minipage}[t]{0.98\linewidth}
        \centering
        \includegraphics[scale=0.5]{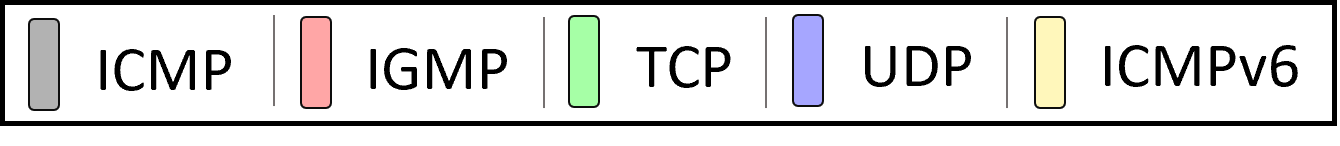}
    \end{minipage}    
    \begin{minipage}[t]{0.98\linewidth}
        \centering
        \input{assets/tikz/ustc/ustc_protocol_dist.tikz} 
        \phantomsubcaption
        \label{app:fig:ustc_proto}
    \end{minipage}
    \begin{minipage}[t]{0.98\linewidth}
        \centering
        \input{assets/tikz/ustc/ustc_unopen_tcp_dist.tikz} 
        \phantomsubcaption
        \label{app:fig:ustc_unopen}
    \end{minipage}    
    \caption{Analysis of USTC2016.}
    \label{fig:ustc_figures}
\end{figure*}

The statistics and characteristics of the USTC dataset are presented in Figure~\ref{fig:ustc_figures}. Fig~\ref{app:fig:ustc_avg_pkts} shows that the average number of packets per unidirectional flow in classes such as BitTorrent, Facetime, Outlook, and Skype is 1 (where Facetime has no downlink packets at all), which results in the inability to calculate flow duration (Fig~\ref{app:fig:ustc_avg_dur_s}) because at least 2 packets are required for such calculations. Similarly in Fig~\ref{app:fig:ustc_avg_iat_ms}, the Inter-Arrival Time statistic couldn't be calculated from many flows that belong to benign classes. The high disparity between the value distribution of benign and malware flows in these figures of time-related features is evident. One can utilize these statistical features in models to differentiate between benign and malware traffic. Furthermore, the protocol distribution in fig~\ref{app:fig:ustc_proto} demonstrates that the benign traffic has quite a low number of different classes that use the UDP protocol, whereas the rest use TCP. Other protocols such as ICMPv6, ICMP, and IGMP can only be found in malware traffic, although in low quantities. Figure~\ref{app:fig:ustc_unopen} shows that compared to benign flows, there is a high number of the malware flows that didn't successfully complete the three-way handshake that is required in the TCP protocol before data can be exchanged between the two endpoints, with one exception of Tinba, which contain UDP-based flows almost exclusively.

\subsection{Datasets Disadvantages}

Any dataset that is utilized for research has drastic effects on the results and conclusions that arise from it, and some issues exist with the publicly available datasets in the wild. The importance of the dataset becomes more highlighted when considering that other components in the ML/DL pipeline depend on it, such as features that are extracted from the data.

\begin{itemize}
    \item Quality: The low quality of the provided network capture files (PCAPs), such as the USTC-TFC2016 dataset, among all the modifications to it such as anonymization, cleaning, and preprocessing that can degrade the data more, may result in inaccurate, non-representative samples of real-life scenarios. 
    While the dataset contains network traffic captures of interesting applications such as MySQL and Facetime, evaluating a model on this dataset may not fully indicate the true prediction performance of it.
    \item Format: Some datasets do not contain raw traffic capture files (PCAPs), but rather a set of features already extracted \cite{QUICClassificationFewLabels} or truncated binary representation of each flow \cite{orange20paper}, which requires special care and non-standard processing. This becomes a problem when trying to compile a dataset from multiple sources. Without the raw files, it is impossible to extract new features with standard tools.
    \item Date: Datasets are quite old, for instance, the ISCXVPN2016, USTC-TFC2016 and Ariel (BOA2016) datasets are from 2016. Still, these datsets are the deafult datasets used in most state-of-the-art works~\cite{Distiller, Wang1DCNN, ODENETFastLeanEncryptedClassification, maldist_ccnc, FlowPic2021, ISCX2016, NetML}\footnote{Other datasets that are used are either lab recordings that are not publicly available, or datasets that are too small for adequate evaluation.}. Applications change over time, along with the increasing trend of adopting encryption with almost every network traffic flow. For example, the network traffic of Skype or Netflix in 2016, might behave differently than the network traffic of Skype or Netflix in 2021.
    \item Scale: There is a bit of disconnection between the research in academia to actual implementation and usage of such innovations in the industry. The datasets feature a lower variety of classes compared to the much bigger scale required in industrialized solutions that sometimes make use of thousands and tens of thousands of different classes. The lack of a sufficient quantity of different classes makes it impossible to evaluate models and features in scenarios that employ a large set of classes in multiple metrics such as the model's performance and training/predicting time.
\end{itemize}

\begin{figure*}[h!]
    \begin{minipage}[t]{0.45\linewidth}    
    \centering
        \includegraphics[width=3.2in]{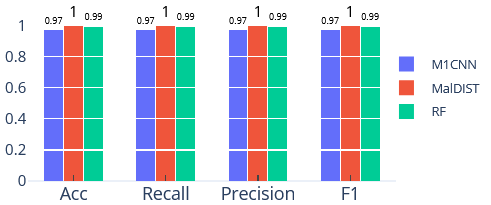}
        \subcaption{Dataset: Ariel $|$ Task: OS Classification.}
        \label{fig:dryrun_ariel_os}
    \end{minipage}
    \hspace{20pt}
    \begin{minipage}[t]{0.45\linewidth}
        \centering
        \includegraphics[width=3.2in]{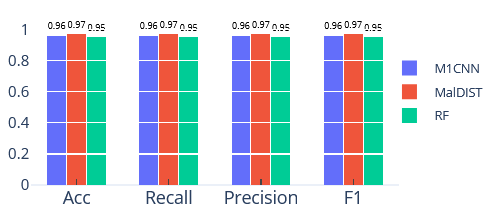}
        \subcaption{Dataset: Ariel $|$ Task: Browser Classification.}
        \label{fig:dryrun_ariel_browser}
    \end{minipage}
    
    \begin{minipage}[t]{0.45\linewidth}
        \centering
        \includegraphics[width=3.2in]{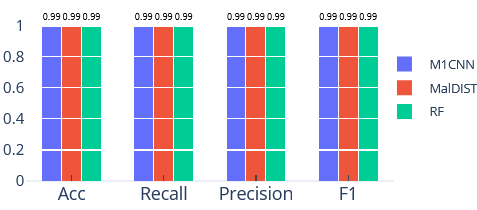}
        \subcaption{Dataset: ISCX2016 $|$ Task: Encapsulation Detection }
        \label{fig:dryrun_iscx_encaps}
    \end{minipage}
    \hspace{20pt}
    \begin{minipage}[t]{0.45\linewidth}
        \centering
        \includegraphics[width=3.2in]{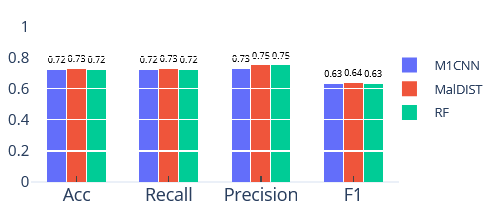}
        \subcaption{Dataset: ISCX2016 $|$ Task: Traffic Type Classification.}
        \label{fig:dryrun_iscx_traffictype}
    \end{minipage}
    
    \begin{minipage}[t]{0.45\linewidth}
        \centering
        \includegraphics[width=3.2in]{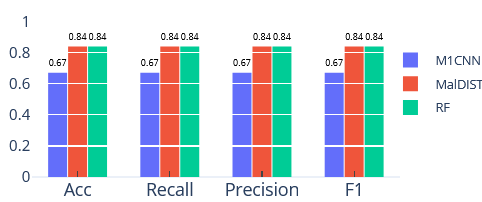}
        \subcaption{Dataset: Selected Flows from ISCX2016 $|$ Task: Encapsulation Detection.}
        \label{fig:dryrun_iscx_selected_encaps}
    \end{minipage}
    \hspace{20pt}
    \begin{minipage}[t]{0.45\linewidth}
        \centering
        \includegraphics[width=3.2in]{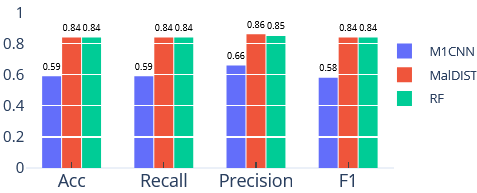}
        \subcaption{Dataset: Selected Flows from ISCX2016 $|$ Task: Traffic Type Classification.}
        \label{fig:dryrun_iscx_selected_traffictype}
    \end{minipage}
    \caption{Framework Evaluation - Same Model on Different Dataset}
    \label{fig:dryrun_results}
\end{figure*}

\begin{figure*}[h!]
    \centering
    \includegraphics[width=6.0in]{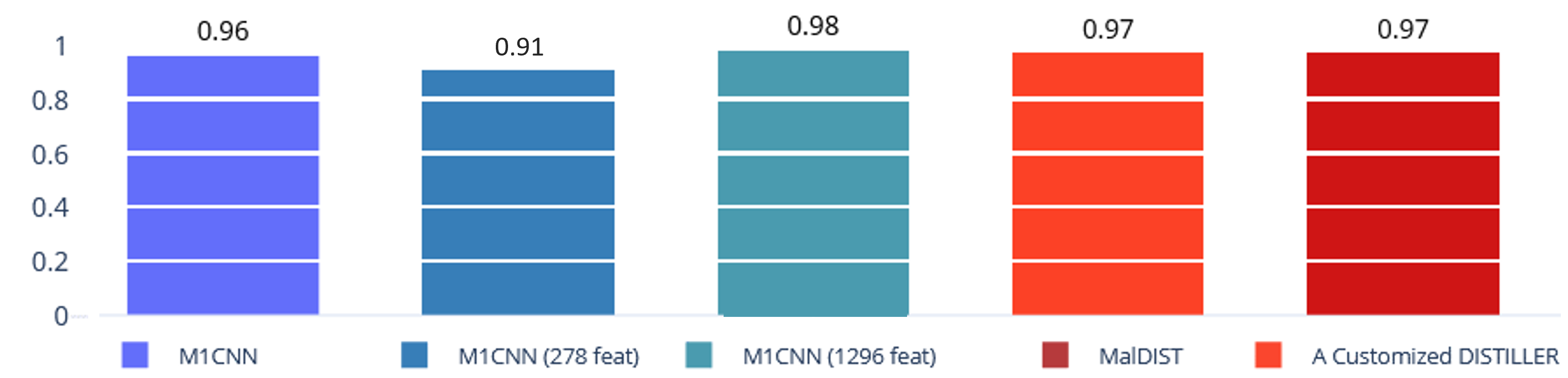}
    \caption{Framework Evaluation - Feature Engineering on the Ariel Dataset for Browser Classification (The Score Values are for all Metrics of Accuracy, Recall, Precision and F1).}
    \label{fig:dryrun_results_new_features}
\end{figure*}

\section{Framework evaluation} \label{sec:dry-run-evaluation}
In this section, we demonstrate how the framework can be used in multiple scenarios: \textbf{I.} We evaluated our framework in the naive case of replicating the results of previous work (Shapira and Shavitt FlowPic~\cite{FlowPic2021}). \textbf{II.} We used a set of three different models (i.e., M1CNN, MalDIST, and RF) on two different datasets (ISCX2016 \cite{ISCX2016_Dataset}, and Ariel \cite{BOA_conf}) to demonstrate the framework's ability to easily transfer the same experimental design (model and features) to a new dataset. \textbf{III.} We provide an example of how one can enhance the feature set of a model, using the plugins provided in our framework. Our main goal, across all three scenarios, is to show that our framework allows one to choose the dataset, the features sets, and the models to train, while replicating previous works, or transferring previous works to new datasets and/or new feature sets and models. The code that was used in this section is publicly available at \cite{DryRunCodeSamples_GitHub}.

As stated above, we began by replicating previous work using our framework. We ran the FlowPic's model on the ISCX2016 \cite{ISCX2016_Dataset} dataset (using the same flows that the authors used in their evaluation) to validate the implementation of both its feature extraction and the model as provided in the framework. We began by executing the learning pipeline, from data to evaluation, that the framework offers. This includes, choosing the dataset, extracting FlowPic images as features from the dataset, selecting the matching deep-learning model, and then evaluating it. Similar to the results reported in \cite{FlowPic2021}, the model achieved over a 95\% accuracy score with our test set on the task of categorizing flows to their traffic type (audio/video/chat and so) over VPN or non-VPN traffic.

We then tested another framework functionality by evaluating a set of models and their respective features on different datasets. The datasets were ISCX2016 \cite{ISCX2016_Dataset} and Ariel~\cite{BOA_conf}. We used the raw datasets without any preprocessing steps before extracting the features to show the applicability of the framework. We also included a third dataset based on selected flows from the ISCX2016 dataset, which were shared with us by the authors of FlowPic \cite{FlowPic2021}, for highly accurate flow-based labeling. From each dataset, we extracted the first 784 payload bytes (using the \textit{n\_bytes} plugin), Lopez-Martin's features of Protocol Headers fields (using the \textit{Protocol Headers} plugin), STNN-inspired features (using the \textit{STNN} plugin), and packet \& TLS record clumps features. In light of the fact that both datasets already use filename-based labeling of the data, we utilized the labeling method based on filenames (which is included in the framework). We selected M1CNN, MalDIST, and Random Forest (RF) as our models for evaluation. For the random forest model, we selected 24 clump-based features of both packets and TLS records.
The results are presented in Fig~\ref{fig:dryrun_results}. Using the Ariel dataset, the models performed superbly with a score of over 95\% for all metrics in both OS and browser classification tasks, perfectly replicating the original results as reported in~\cite{BOA_conf}. On the other hand, the models demonstrated several difficulties classifying flows by their encapsulation (VPN or non-VPN) and traffic type (e.g., video, audio, chat, or file transfer) with the ISCX2016 dataset. In this run, the framework alleviated the process of evaluating models over different datasets. While the deep-learning models, such as M1CNN and MalDIST, were fed their proposed features (extracted using their respective plugins), we carefully selected 24 features for the Random Forest classifier.

\begin{figure*}[h!]
    \begin{minipage}[t]{0.48\linewidth}
        \centering
        \includegraphics[width=1.2in]{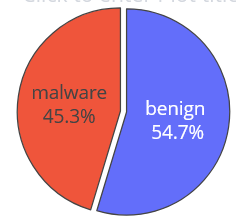}
        \subcaption{Challenge 1 (Detection): Distribution of classes (2 classes) in \textit{train} and \textit{test} sets.}
    \end{minipage}
    \begin{minipage}[t]{0.48\linewidth}
        \centering
        \includegraphics[width=1.15in]{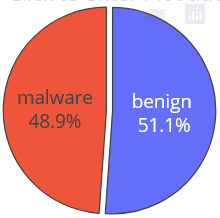}
        \subcaption{Challenge 3 (Zero-day): Distribution of classes (2 classes) in \textit{train} (known traffic) and \textit{test} (unknown traffic),}
    \end{minipage}
    \hspace{20pt}
    \begin{minipage}[t]{0.97\linewidth}
        \centering
        \includegraphics[width=5.14in]{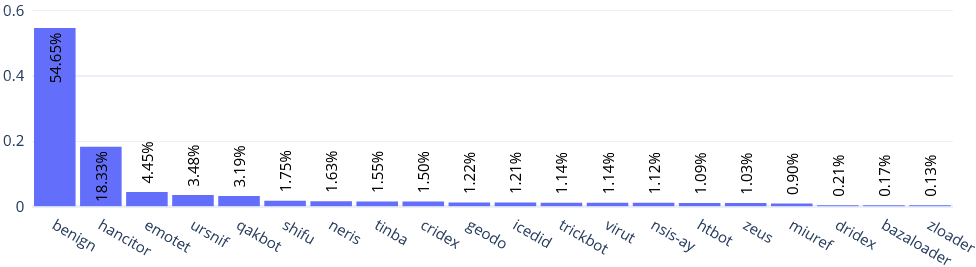}
        \subcaption{Challenge 2 (Classification): Distribution of classes (20 in total) in \textit{train} and \textit{test} sets.}
    \end{minipage}
    \caption{Sample Distribution per Class in the Challenges.}
    \label{fig:challenge-sample-dist}
\end{figure*}

One of the main parts of the learning pipeline is the feature engineering phase. Therefore, for the last scenario, we demonstrate the framework plugins ability, by feeding the deep-learning architecture with different input sizes and features. For M1CNN, we used an input size of 278 features, where the first 200 features were as suggested by the authors (i.e., the first 200 payload bytes of the flow out of the suggested 784 payload bytes), 70 features of a flattened array of STNN-inspired features, and a total of 8 other crafted statistical features (generated by the \textit{PacketRelativeTime}, \textit{SmallPacketPayloadRatio}, and \textit{ResReqDiffTime} plugins). 

We repeated the experiment of the second scenario with several changes. We fed the M1CNN model with 1296 features comprising the 784 payload bytes as suggested by the authors, along with an additional 512 features of byte frequency in the first 6 packets (using the \textit{NPacketsByteFrequency} plugin), namely 256 features for each direction of the flow (byte values are in [0,255]). 

WLOG, we evaluated the model on the Ariel dataset, choosing the browser classification task. The M1CNN (278 feat) model achieved a result of 91\%  for all metrics as depicted in Figure \ref{fig:dryrun_results_new_features}, which is a bit lower than the results yielded by M1CNN with the default input size and features with which we experimented as detailed in the previous paragraph (i.e., 96\% for all metrics, see Fig~\ref{fig:dryrun_results}). Figure  \ref{fig:dryrun_results_new_features} demonstrate that in the case of M1CNN (1296 feat), the evaluation of the model resulted in 98\% for all metrics, surpassing all others.

Intrigued by the results of this M1CNN (1296 feat), we decided to use the DISTILLER architecture with customized plug-in components, in terms of modalities (inputs) and tasks (outputs). Therefore, we built a variant of DISTILLER which contained two modalities and two tasks (OS and browser). The first modality took the first 784 payload bytes of a flow as input, while the second modality took the byte frequencies (the same features that were fed into our last M1CNN) as input. This custom DISTILLER variant achieved a 97\% score for all metrics for the task of browser classification. as can be seen in Fig. Fig~\ref{fig:dryrun_results_new_features}. The figure shows that our framework provides the ability to use new features over well-known models. Note that in some cases this can improve the results (e.g., M1CNN - 1296 feat), while in other cases (e.g., M1CNN - 278 feat) it may decrease them. We encourage researchers to experiment with the feature engineering ability of the framework, which is simple and easy, while being aware that not every transformation is beneficial at the end of the road. 

\section{Online Challenges} \label{sec:online-challenge-evalai}

In this section, we present online challenges hosted by EvalAI \cite{EvalAI, evalai_website}. EvalAI is an open-source platform for evaluating and comparing machine-learning models at scale. We created three challenges in the malware traffic detection and classification domain.

The three challenges are as follows:
\begin{enumerate}
    \item Detection of malware traffic (binary): Distinguish between benign and malware traffic.
    \item Classification of malware traffic (multi-class): Classify known malware family traffic.
    \item Zero-day detection (binary): Detect unknown malware family traffic.
\end{enumerate}

The public datasets that we utilized include:
\begin{itemize}
    \item Benign: ISCX2016 \cite{ISCX2016_Dataset}, StratosphereIPS \cite{stratodatasets}, and the benign subset of USTC2016 \cite{USTC}.
    \item Malware: MTA \cite{MTA} and the malware subset of USTC2016 \cite{USTC}.
\end{itemize}

For each challenge, we provide the PCAP files for the \textit{train} and \textit{test} sets. Due to the class imbalance in the test sets of the first two challenges, we decided to use F1-score as the leading metric that the leader-board will follow to determine the best results. For the third challenge, zero-day detection, we decided to use the metrics of the detection rate, indicated by the True Positive Rate (TPR) and the False Alarm Rate (FAR), which is also known as the false positive rate. The leading metric in this challenge is a combination of the two: 
\begin{equation}
        TPR\cdot (1-FAR)    
\end{equation}

Where the TPR and FAR values are calculated by the equations of:
\begin{equation}
\begin{split}
    TPR & = \frac{TP}{TP+FN} \\
    FAR & = \frac{FP}{TN+FP}
\end{split}
\end{equation}

The description of the detailed challenges can be found on the challenges page on GitHub: \textit{https://github.com/ArielCyber/OSF-EIMTC-Challenge}. We encourage the participants to use our framework \cite{OSFEIMTC_GitHub} while attempting to tackle the presented challenges.



\section{Summary} \label{sec:summary}
The problem of classifying encrypted traffic will become more popular and important once the adoption of privacy-concerned and encrypted protocols such as DoH and QUIC will increase and gain momentum. This calls for extensive research and collaboration by developing and extending the framework with full ML/DL pipelines to tackle new protocols and other tasks in various domains. As a result, this paper presents a full pipeline framework. The framework creates a soft landing for new researchers to enter the domain of traffic classification. By providing organized access to datasets, feature extraction, and implementations for state-of-the-art deep-learning models, The framework allows to researchers to easily compare many models and feature sets, generating a rich comparison of a variety of solutions. 

Researchers can also extend our framework to support new scopes of features, such as time windows and host-based features. By contributing to the framework, with new plugins of feature sets or implementations of newly proposed models, the research community will be able to save tremendous time when evaluating processes and models, facilitate more accurate comparisons, and enable reproducible experiments. New quality datasets with a vast number of classes and standardized formats (e.g., PCAP and PCAPNG), especially for new and upcoming protocols such as QUIC and TLS 1.3. are required to boost the research quality and the quality of the evaluations of proposed solutions. Any researcher can use the same plugins and tools that are provided in the proposed framework, to easily make use of new quality datasets. We expect that the online challenges that we have described in section \ref{sec:online-challenge-evalai} will promote the collaboration of the research community by contributing to the framework.


\bibliographystyle{IEEEtran}
\bibliography{ref}

\begin{appendices}
\section{Ariel Dataset Analysis}
\begin{figure*}[h!]
    \centering

    \begin{minipage}[t]{0.95\linewidth}
        \centering
        \includegraphics[scale=0.5]{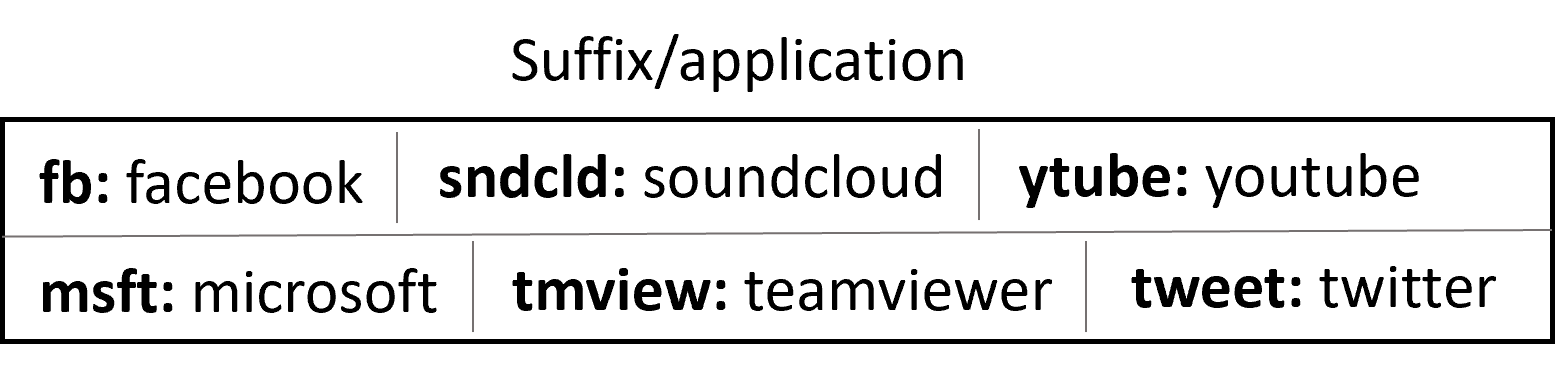}
    \end{minipage}
    \begin{minipage}[t]{0.52\linewidth}
        \hspace*{0.29\linewidth}
       \includegraphics[scale=0.5]{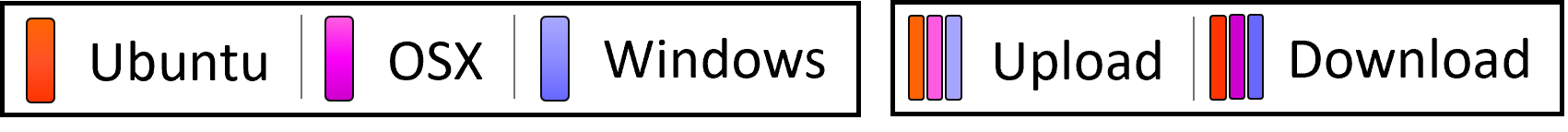}
    \end{minipage}
    \begin{minipage}[t]{0.46\linewidth}
        \includegraphics[scale=0.5]{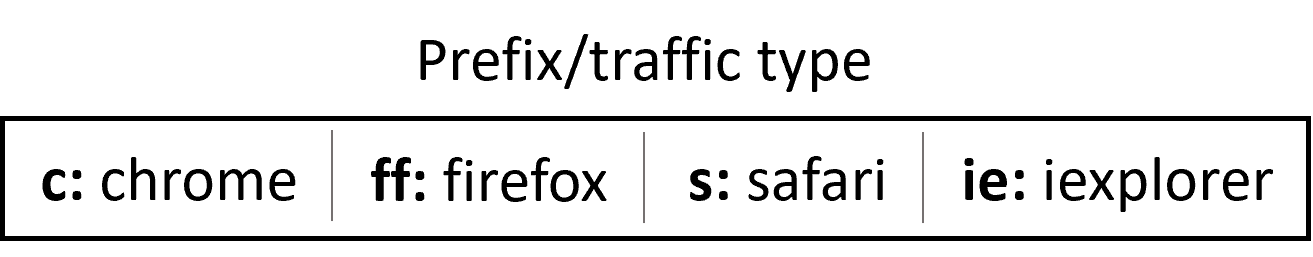}
    \end{minipage}
    
    \begin{minipage}[t]{0.75\linewidth}
        \centering
        \input{assets/tikz/boa/boa_avg_packet_size.tikz}
        \phantomsubcaption
        \label{app:fig:boa_avg_pkt_size}
    \end{minipage}
    \begin{minipage}[t]{0.75\linewidth}
        \centering
          \input{assets/tikz/boa/boa_avg_packets.tikz}
        \phantomsubcaption
        \label{app:fig:boa_avg_pkts}
    \end{minipage}
    \begin{minipage}[t]{0.75\linewidth}
        \centering
        \input{assets/tikz/boa/boa_avg_bytes.tikz}
        \phantomsubcaption
        \label{app:fig:boa_avg_bytes}
    \end{minipage}
    \begin{minipage}[t]{0.75\linewidth}
        \centering
        \input{assets/tikz/boa/boa_avg_duration_s.tikz}
        \phantomsubcaption
        \label{app:fig:boa_avg_dur_s}
    \end{minipage}
    \begin{minipage}[t]{0.75\linewidth}
        \centering
       \input{assets/tikz/boa/boa_avg_iat_ms.tikz}
        \phantomsubcaption
        \label{app:fig:boa_avg_iat_ms}
    \end{minipage}
    \caption{Analysis of Ariel (BOA2016).}
    \label{app:fig:boa}
\end{figure*}

The figures for this dataset are depicted in figure~\ref{app:fig:boa}. It is apparent that not all browsers share the same set of applications. For example, \textit{Facebook} (fb) is only shared among \textit{Chrome} and \textit{Firefox} labels. Furthermore, \textit{Ubuntu} is the only operating system that hosted that combination of browser and application. A similar observation on the different operating systems (colored in bars) is that some applications and browsers can only be found in the data among specific operating systems, some for technical reasons such as \textit{Internet Explorer} and \textit{Safari}, which are exclusively for \textit{Windows} and \textit{OSX} operating systems respectively. From figure~\ref{app:fig:boa_avg_pkt_size} it is evident that across all labels, with the exception of \textit{c-tweet} (Windows), the packet size was larger on average in the download flow than the upload flow, This trend can also be seen in figures \ref{app:fig:boa_avg_pkts} and \ref{app:fig:boa_avg_bytes} illustrating the average number of packets and bytes per session. Note that there are some non-web applications that operated in the background while recording, which are \textit{Dropbox}, \textit{Microsoft} (services), \textit{Vine}, and \textit{Teamviewer}, hence, they do not contain a browser label.

\section{ISCX2016 Dataset Analysis}
\begin{figure*}[h!]
    \begin{minipage}[t]{0.95\linewidth}
        \centering
        \includegraphics[scale=0.5]{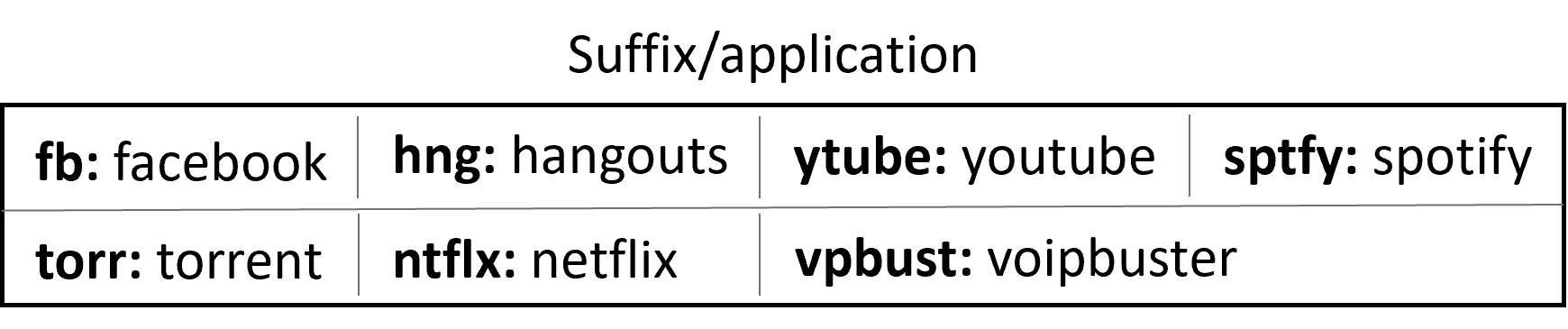}
    \end{minipage}
    \begin{minipage}[t]{0.48\linewidth}
        \hspace*{0.35\linewidth}
       \includegraphics[scale=0.5]{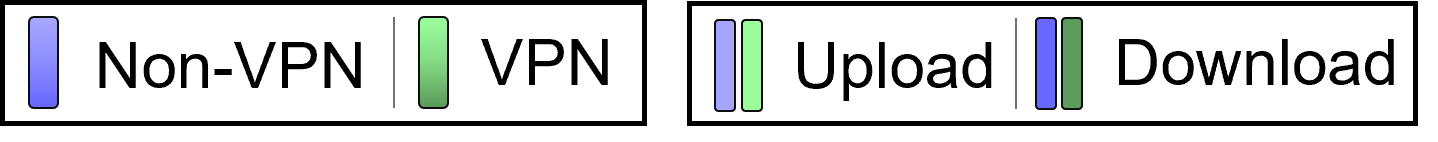}
    \end{minipage}
        \begin{minipage}[t]{0.48\linewidth}
        \includegraphics[scale=0.5]{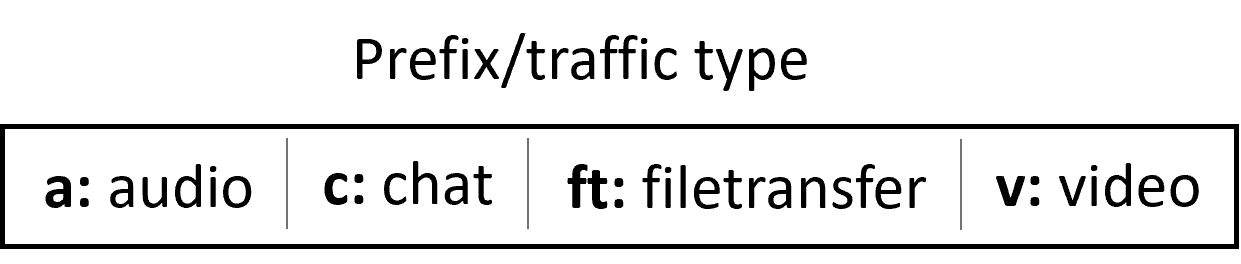}
    \end{minipage}
    
    \begin{minipage}[t]{0.95\linewidth}
        \centering
        \input{assets/tikz/iscx/iscx_avg_packet_size.tikz}
        \phantomsubcaption
        \label{app:fig:iscx_avg_pkt_size}
    \end{minipage}
    \begin{minipage}[t]{0.95\linewidth}
        \centering
          \input{assets/tikz/iscx/iscx_avg_packets.tikz}
        \phantomsubcaption
        \label{app:fig:iscx_avg_pkts}
    \end{minipage}
    \begin{minipage}[t]{0.95\linewidth}
        \centering
        \input{assets/tikz/iscx/iscx_avg_bytes.tikz}
        \phantomsubcaption
        \label{app:fig:iscx_avg_bytes}
    \end{minipage}
    \begin{minipage}[t]{0.95\linewidth}
        \centering
        \input{assets/tikz/iscx/iscx_avg_duration_s.tikz}
        \phantomsubcaption
        \label{app:fig:iscx_avg_dur_s}
    \end{minipage}
    \begin{minipage}[t]{0.95\linewidth}
        \centering
        \input{assets/tikz/iscx/iscx_avg_iat_ms.tikz}
        \phantomsubcaption
        \label{app:fig:iscx_avg_iat_ms}
    \end{minipage}
    
    \begin{minipage}[t]{0.95\linewidth}
        \centering
        \includegraphics[scale=0.5]{assets/tikz/ustc/ustc_legend_proto.png}
    \end{minipage} 
    \begin{minipage}[t]{0.95\linewidth}
        \centering
        \input{assets/tikz/iscx/iscx_protocol_dist_nonvpn.tikz} 
        \phantomsubcaption
        \label{app:fig:iscx_proto_nonvpn}
    \end{minipage}    
    \begin{minipage}[t]{0.95\linewidth}
        \centering
        \input{assets/tikz/iscx/iscx_protocol_dist.tikz} 
        \phantomsubcaption
        \label{app:fig:iscx_proto}
    \end{minipage}  
    \caption{Analysis of ISCX2016.}
    \label{app:fig:iscx}
\end{figure*}

The figures for this dataset are depicted in figure~\ref{app:fig:iscx}. While most of the applications can be found in both \textit{VPN} and \textit{non-VPN} traffic, there is a relatively small number of applications that are exclusive to a one category of encapsulation (only \textit{non-VPN}). Namely \textit{Gmail}, \textit{scp}, \textit{Facebook}, and the video portions of \textit{Hangouts} and \textit{Skype}. In a swift glance at the error lines in figures \ref{app:fig:iscx_avg_pkts}, \ref{app:fig:iscx_avg_bytes}, \ref{app:fig:iscx_avg_dur_s}, and \ref{app:fig:iscx_avg_iat_ms}, it is evident that the standard deviation is very high for the number of packets, bytes, duration, and inter-arrival time per session, hinting that each feature on its own might not be a most excellent separator between the classes. From the two figures of protocol frequencies per label, for \textit{non-VPN} \ref{app:fig:iscx_proto_nonvpn} and for VPN \ref{app:fig:iscx_proto}, the sessions with protocols other than UDP and TCP are more apparent in \textit{non-VPN} than \textit{VPN} encapsulated traffic. Both in frequency and in the set of different protocols. A researcher that adopt this dataset might want to clean the less important protocols while attempting to classify a network flow to its traffic type and application.

\section{MAppGraph Dataset portion Analysis}
\begin{figure*}[h!]
    \begin{minipage}[t]{0.95\linewidth}
        \hspace*{0.41\linewidth}
       \includegraphics[scale=0.5]{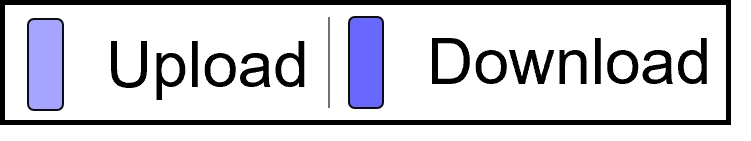}
    \end{minipage}
    
    \begin{minipage}[t]{0.95\linewidth}
        \centering
        \input{assets/tikz/mappgraph/mg_avg_packet_size.tikz}
        \phantomsubcaption
        \label{app:fig:mg_avg_pkt_size}
    \end{minipage}
    \begin{minipage}[t]{0.95\linewidth}
        \centering
          \input{assets/tikz/mappgraph/mg_avg_packets.tikz}
        \phantomsubcaption
        \label{app:fig:mg_avg_pkts}
    \end{minipage}
    \begin{minipage}[t]{0.95\linewidth}
        \centering
        \input{assets/tikz/mappgraph/mg_avg_bytes.tikz}
        \phantomsubcaption
        \label{app:fig:mg_avg_bytes}
    \end{minipage}
    \begin{minipage}[t]{0.95\linewidth}
        \centering
        \input{assets/tikz/mappgraph/mg_avg_duration_s.tikz}
        \phantomsubcaption
        \label{app:fig:mg_avg_dur_s}
    \end{minipage}
    \begin{minipage}[t]{0.95\linewidth}
        \centering
        \input{assets/tikz/mappgraph/mg_avg_iat_ms.tikz}
        \phantomsubcaption
        \label{app:fig:mg_avg_iat_ms}
    \end{minipage}
    
    \begin{minipage}[t]{0.95\linewidth}
        \centering
        \includegraphics[scale=0.5]{assets/tikz/ustc/ustc_legend_proto.png}
    \end{minipage} 
    \begin{minipage}[t]{0.95\linewidth}
        \centering
        \input{assets/tikz/mappgraph/mg_protocol_dist.tikz} 
        \phantomsubcaption
        \label{app:fig:mg_proto}
    \end{minipage}    
    \begin{minipage}[t]{0.95\linewidth}
        \centering
        \input{assets/tikz/mappgraph/mg_unopen_tcp_dist.tikz}
        \phantomsubcaption
        \label{app:fig:mg_unopen}
    \end{minipage}    
    \caption{Analysis of a portion of MAppGraph dataset (of 8 applications).}
    \label{app:fig:mg}
\end{figure*}

The figures for this dataset are depicted in figure~\ref{app:fig:mg}. The flow-based protocol distribution of TCP and UDP (in figure~\ref{app:fig:mg_proto} is almost even across each application, there is no application that is drastically more UDP or TCP based. Which is unique to this dataset portion when compared to the other datasets we have analyzed. In all graphs, except for average packet size (fig~\ref{app:fig:mg_avg_pkt_size}), it is prevalent that the standard deviation is high, which might be related to the difference in statistical characteristics between UDP and TCP sessions. From figure \ref{app:fig:mg_unopen} we see that \textit{Skype} has the most unopened TCP sessions with 159 sessions, which amounts to 27.6\% of the total TCP sessions of \textit{Skype}. While on the other end there is \textit{Soundcloud} with a single unopened TCP session.

\end{appendices}

\end{document}

%% file: assets/tikz/ustc/ustc_avg_packet_size.tikz
\begin{tikzpicture}
 
\begin{axis} [ybar=0.1pt,
height = 4cm, width = 12.5cm,
bar width = 4pt,
title=\textcolor{blue}{(\ref{app:fig:ustc_avg_pkt_size})} AVG Pkt Size x Session,every axis title/.style={below right,at={(0,1)},draw=gray,fill=black!5,font=\fontsize{6}{6}\selectfont},
ymin = 1,
ymax = 10000,
ymode=log,
label style={font=\small},
log basis y={10},
ytick distance=10,
ymajorgrids=true,
xticklabel shift={-5pt},
yticklabel shift={-2pt},
enlarge x limits=0.025,
symbolic x coords={
torrent,ftp,mysql,weibo,smb,gmail,wow,outlook,skype,facetime,geodo,neris,virut,cridex,miuref,zeus,shifu,htbot,nsis-ay,tinba
},
xtick={
ftp,mysql,geodo,weibo,smb,neris,virut,cridex,miuref,zeus,shifu,gmail,tinba,wow,outlook,torrent,htbot,skype,nsis-ay,facetime
},
x tick label style={
font=\fontsize{3.15}{1}\selectfont
},
y tick label style={font=\tiny},
xtick style={draw=none},
ytick style={draw=none},
]

\addplot+[
    blue!90!black,fill=blue!35!white,
    bar shift=-2.1pt,
    error bars/.cd,
    y dir=both,
    y explicit relative,
    error mark options={
      rotate=90,
      mark size=0.9pt,
      line width=0.5pt
    }
] coordinates {
    (facetime,403.841)       +- (0, 0.646)
    (ftp,213.663)            +- (0, 1.464)
    (gmail,565.271)          +- (0, 0.824)
    (mysql,121.778)          +- (0, 0.958)
    (outlook,1452.335)       +- (0, 0.114)
    (skype,608.199)          +- (0, 0.317)
    (smb,1479.016)           +- (0, 0.099)
    (torrent,920.920)        +- (0, 0.652)
    (weibo,1494.292)         +- (0, 0.016)
    (wow,84.202)             +- (0, 0.022)
};
\addplot+[
    blue!90!black,fill=blue!60!white,
    bar shift=2.1pt,
    error bars/.cd,
    y dir=both,
    y explicit relative,
    error mark options={
      rotate=90,
      mark size=0.9pt,
      line width=0.5pt
    }
] coordinates {
    (facetime,0.000)         +- (0, 0.000)
    (ftp,89.485)             +- (0, 1.038)
    (gmail,94.627)           +- (0, 0.759)
    (mysql,77.628)           +- (0, 0.184)
    (outlook,70.006)         +- (0, 0.002)
    (skype,70.000)           +- (0, 0.000)
    (smb,72.106)             +- (0, 0.543)
    (torrent,70.000)         +- (0, 0.000)
    (weibo,70.471)           +- (0, 0.364)
    (wow,107.132)            +- (0, 0.019)
};

\addplot+[
    red!90!black,fill=red!35!white,
    bar shift=-2.1pt,
    error bars/.cd,
    y dir=both,
    y explicit relative,
    error mark options={
      rotate=90,
      mark size=0.9pt,
      line width=0.5pt
    }
] coordinates{
    (cridex,154.646)         +- (0, 0.405)
    (geodo,96.919)           +- (0, 0.381)
    (htbot,497.631)          +- (0, 0.947)
    (miuref,199.300)         +- (0, 0.673)
    (neris,119.746)          +- (0, 1.752)
    (nsis-ay,117.555)        +- (0, 1.236)
    (shifu,104.659)          +- (0, 0.146)
    (tinba,79.610)           +- (0, 0.124)
    (virut,196.131)          +- (0, 1.646)
    (zeus,87.995)            +- (0, 0.713)
};

\addplot+[
    red!90!black,fill=red!60!white,
    bar shift=2.1pt,
    error bars/.cd,
    y dir=both,
    y explicit relative,
    error mark options={
      rotate=90,
      mark size=0.9pt,
      line width=0.5pt
    }
] coordinates{
    (cridex,269.185)         +- (0, 0.019)
    (geodo,226.392)          +- (0, 1.429)
    (htbot,504.192)          +- (0, 0.928)
    (miuref,167.409)         +- (0, 0.541)
    (neris,302.907)          +- (0, 1.774)
    (nsis-ay,1110.702)       +- (0, 0.415)
    (shifu,151.966)          +- (0, 0.752)
    (tinba,140.988)          +- (0, 0.005)
    (virut,363.432)          +- (0, 1.917)
    (zeus,266.572)           +- (0, 1.760)
};
\end{axis}
\end{tikzpicture}

%% file: assets/tikz/ustc/ustc_avg_packets.tikz
\begin{tikzpicture}
 
\begin{axis} [ybar=0.1pt,
height = 4cm, width = 12.5cm,
bar width = 4pt,
title=\textcolor{blue}{(\ref{app:fig:ustc_avg_pkts})} AVG Pkts x Session,every axis title/.style={below right,at={(0,1)},draw=gray,fill=black!5,font=\fontsize{6}{6}\selectfont},
ymin = 1,
ymode=log,
label style={font=\small},
log basis y={10},
ytick distance=10,
ymajorgrids=true,
xticklabel shift={-5pt},
yticklabel shift={-2pt},
enlarge x limits=0.025,
symbolic x coords={
torrent,ftp,mysql,weibo,smb,gmail,wow,outlook,skype,facetime,geodo,neris,virut,cridex,miuref,zeus,shifu,htbot,nsis-ay,tinba
},
xtick={
ftp,mysql,geodo,weibo,smb,neris,virut,cridex,miuref,zeus,shifu,gmail,tinba,wow,outlook,torrent,htbot,skype,nsis-ay,facetime
},
x tick label style={
font=\fontsize{3.15}{1}\selectfont
},
y tick label style={font=\tiny},
xtick style={draw=none},
ytick style={draw=none},
]

\addplot+[
    blue!90!black,fill=blue!35!white,
    bar shift=-2.1pt,
    error bars/.cd,
    y dir=both,
    y explicit relative,
    error mark options={
      rotate=90,
      mark size=0.9pt,
      line width=0.5pt
    }
] coordinates {
    (facetime,1.000)         +- (0, 0.000)
    (ftp,2.008)              +- (0, 0.650)
    (gmail,1.658)            +- (0, 0.464)
    (mysql,1.243)            +- (0, 0.473)
    (outlook,1.002)          +- (0, 0.047)
    (skype,1.000)            +- (0, 0.000)
    (smb,22.262)             +- (0, 0.571)
    (torrent,1.000)          +- (0, 0.000)
    (weibo,28.652)           +- (0, 0.272)
    (wow,8.888)              +- (0, 0.086)    
};
\addplot+[
    blue!90!black,fill=blue!60!white,
    bar shift=2.1pt,
    error bars/.cd,
    y dir=both,
    y explicit relative,
    error mark options={
      rotate=90,
      mark size=0.9pt,
      line width=0.5pt
    }
] coordinates {
    (facetime,0.000)         +- (0, 0.000)
    (ftp,1.555)              +- (0, 0.668)
    (gmail,1.240)            +- (0, 0.534)
    (mysql,1.080)            +- (0, 0.254)
    (outlook,0.992)          +- (0, 0.093)
    (skype,0.898)            +- (0, 0.336)
    (smb,1.506)              +- (0, 0.680)
    (torrent,0.995)          +- (0, 0.067)
    (weibo,1.637)            +- (0, 0.726)
    (wow,8.871)              +- (0, 0.098)
};

\addplot+[
    red!90!black,fill=red!35!white,
    bar shift=-2.1pt,
    error bars/.cd,
    y dir=both,
    y explicit relative,
    error mark options={
      rotate=90,
      mark size=0.9pt,
      line width=0.5pt
    }
] coordinates{
    (cridex,17.181)          +- (0, 5.217)
    (geodo,4.441)            +- (0, 50.063)
    (htbot,13.356)           +- (0, 7.574)
    (miuref,4.362)           +- (0, 16.722)
    (neris,10.366)           +- (0, 26.837)
    (nsis-ay,16.533)         +- (0, 9.854)
    (shifu,50.884)           +- (0, 66.030)
    (tinba,1.464)            +- (0, 10.981)
    (virut,9.320)            +- (0, 2.777)
    (zeus,5.444)             +- (0, 5.846)
};

\addplot+[
    red!90!black,fill=red!60!white,
    bar shift=2.1pt,
    error bars/.cd,
    y dir=both,
    y explicit relative,
    error mark options={
      rotate=90,
      mark size=0.9pt,
      line width=0.5pt
    }
] coordinates{
    (cridex,10.976)          +- (0, 1.018)
    (geodo,0.766)            +- (0, 5.904)
    (htbot,13.225)           +- (0, 6.117)
    (miuref,1.659)           +- (0, 1.469)
    (neris,4.324)            +- (0, 37.537)
    (nsis-ay,40.488)         +- (0, 6.540)
    (shifu,0.961)            +- (0, 1.668)
    (tinba,1.111)            +- (0, 0.536)
    (virut,3.881)            +- (0, 3.424)
    (zeus,2.412)             +- (0, 5.624)
};

\end{axis}

\end{tikzpicture}

%% file: assets/tikz/ustc/ustc_avg_bytes.tikz
\begin{tikzpicture}
 
\begin{axis} [ybar=0.1pt,
height = 4cm, width = 12.5cm,
bar width = 4pt,
title=\textcolor{blue}{(\ref{app:fig:ustc_avg_bytes})} AVG Bytes x Session,every axis title/.style={below right,at={(0,1)},draw=gray,fill=black!5,font=\fontsize{6}{6}\selectfont},
ymin = 1,
ymode=log,
label style={font=\small},
log basis y={10},
ytick distance=10,
ymajorgrids=true,
xticklabel shift={-5pt},
yticklabel shift={-2pt},
enlarge x limits=0.025,
symbolic x coords={
torrent,ftp,mysql,weibo,smb,gmail,wow,outlook,skype,facetime,geodo,neris,virut,cridex,miuref,zeus,shifu,htbot,nsis-ay,tinba
},
xtick={
ftp,mysql,geodo,weibo,smb,neris,virut,cridex,miuref,zeus,shifu,gmail,tinba,wow,outlook,torrent,htbot,skype,nsis-ay,facetime
},
x tick label style={
font=\fontsize{3.15}{1}\selectfont
},
y tick label style={font=\tiny},
xtick style={draw=none},
ytick style={draw=none},
]

\addplot+[
    blue!90!black,fill=blue!35!white,
    bar shift=-2.1pt,
    error bars/.cd,
    y dir=both,
    y explicit relative,
    error mark options={
      rotate=90,
      mark size=0.9pt,
      line width=0.5pt
    }
] coordinates {
    (facetime,403.841)       +- (0, 0.646)
    (ftp,429.007)            +- (0, 2.346)
    (gmail,936.965)          +- (0, 0.603)
    (mysql,151.382)          +- (0, 0.827)
    (outlook,1455.230)       +- (0, 0.106)
    (skype,608.199)          +- (0, 0.317)
    (smb,32925.379)          +- (0, 0.594)
    (torrent,920.920)        +- (0, 0.652)
    (weibo,42814.244)        +- (0, 0.273)
    (wow,748.422)            +- (0, 0.081)   
};
\addplot+[
    blue!90!black,fill=blue!60!white,
    bar shift=2.1pt,
    error bars/.cd,
    y dir=both,
    y explicit relative,
    error mark options={
      rotate=90,
      mark size=0.9pt,
      line width=0.5pt
    }
] coordinates {
    (facetime,0.000)         +- (0, 0.000)
    (ftp,139.166)            +- (0, 2.057)
    (gmail,117.305)          +- (0, 1.633)
    (mysql,83.845)           +- (0, 0.464)
    (outlook,69.420)         +- (0, 0.093)
    (skype,62.890)           +- (0, 0.336)
    (smb,108.609)            +- (0, 2.380)
    (torrent,69.683)         +- (0, 0.067)
    (weibo,115.393)          +- (0, 1.570)
    (wow,950.407)            +- (0, 0.104)
};

\addplot+[
    red!90!black,fill=red!35!white,
    bar shift=-2.1pt,
    error bars/.cd,
    y dir=both,
    y explicit relative,
    error mark options={
      rotate=90,
      mark size=0.9pt,
      line width=0.5pt
    }
] coordinates{
    (cridex,2656.898)        +- (0, 4.965)
    (geodo,430.433)          +- (0, 75.416)
    (htbot,6646.332)         +- (0, 11.300)
    (miuref,869.298)         +- (0, 12.318)
    (neris,1241.303)         +- (0, 16.590)
    (nsis-ay,1943.590)       +- (0, 10.719)
    (shifu,5325.434)         +- (0, 67.549)
    (tinba,116.576)          +- (0, 14.391)
    (virut,1827.861)         +- (0, 8.709)
    (zeus,479.040)           +- (0, 13.298)
};

\addplot+[
    red!90!black,fill=red!60!white,
    bar shift=2.1pt,
    error bars/.cd,
    y dir=both,
    y explicit relative,
    error mark options={
      rotate=90,
      mark size=0.9pt,
      line width=0.5pt
    }
] coordinates{
    (cridex,2954.481)        +- (0, 1.018)
    (geodo,173.387)          +- (0, 30.382)
    (htbot,6667.726)         +- (0, 10.587)
    (miuref,277.807)         +- (0, 1.730)
    (neris,1309.856)         +- (0, 13.068)
    (nsis-ay,44970.012)      +- (0, 7.305)
    (shifu,145.976)          +- (0, 5.428)
    (tinba,156.613)          +- (0, 0.536)
    (virut,1410.363)         +- (0, 7.939)
    (zeus,643.032)           +- (0, 31.578)
};

\end{axis}

\end{tikzpicture}

%% file: assets/tikz/ustc/ustc_avg_duration_s.tikz
\begin{tikzpicture}
 
\begin{axis} [ybar=0.1pt,
height = 4cm, width = 12.5cm,
bar width = 4pt,
title=\textcolor{blue}{(\ref{app:fig:ustc_avg_dur_s})} AVG Duration (sec) x Session,every axis title/.style={below right,at={(0,1)},draw=gray,fill=black!5,font=\fontsize{6}{6}\selectfont},
ymin = 1,
ymode=log,
label style={font=\small},
log basis y={10},
ytick distance=10,
ymajorgrids=true,
xticklabel shift={-5pt},
yticklabel shift={-2pt},
enlarge x limits=0.025,
symbolic x coords={
torrent,ftp,mysql,weibo,smb,gmail,wow,outlook,skype,facetime,geodo,neris,virut,cridex,miuref,zeus,shifu,htbot,nsis-ay,tinba
},
xtick={
torrent,ftp,mysql,weibo,smb,gmail,wow,outlook,skype,facetime,geodo,neris,virut,cridex,miuref,zeus,shifu,htbot,nsis-ay,tinba
},
x tick label style={
font=\fontsize{3.15}{1}\selectfont
},
y tick label style={font=\tiny},
xtick style={draw=none},
ytick style={draw=none},
]

\addplot+[
    blue!90!black,fill=blue!35!white,
    bar shift=-2.1pt,
    error bars/.cd,
    y dir=both,
    y explicit relative,
    error mark options={
      rotate=90,
      mark size=0.9pt,
      line width=0.5pt
    }
] coordinates {
    (facetime,0.000)         +- (0, 0.000)
    (ftp,0.021)              +- (0, 1.971)
    (gmail,0.000)            +- (0, 4.709)
    (mysql,0.000)            +- (0, 7.132)
    (outlook,0.000)          +- (0, 43.362)
    (skype,0.000)            +- (0, 0.000)
    (smb,0.002)              +- (0, 8.233)
    (torrent,0.00001)          +- (0, 0.000)
    (weibo,0.001)            +- (0, 11.395)
    (wow,0.001)              +- (0, 0.151)
};
\addplot+[
    blue!90!black,fill=blue!60!white,
    bar shift=2.1pt,
    error bars/.cd,
    y dir=both,
    y explicit relative,
    error mark options={
      rotate=90,
      mark size=0.9pt,
      line width=0.5pt
    }
] coordinates {
    (facetime,0.000)         +- (0, 0.000)
    (ftp,0.021)              +- (0, 1.973)
    (gmail,0.0003)            +- (0, 5.419)
    (mysql,0.000009)            +- (0, 10.696)
    (outlook,0.000)          +- (0, 0.000)
    (skype,0.000)            +- (0, 0.000)
    (smb,0.001)              +- (0, 8.725)
    (torrent,0.000)          +- (0, 0.000)
    (weibo,0.0004)            +- (0, 15.601)
    (wow,0.001)              +- (0, 0.155)

};

\addplot+[
    red!90!black,fill=red!35!white,
    bar shift=-2.1pt,
    error bars/.cd,
    y dir=both,
    y explicit relative,
    error mark options={
      rotate=90,
      mark size=0.9pt,
      line width=0.5pt
    }
] coordinates{
    (cridex,508409.774)      +- (0, 0.178)
    (geodo,22910.379)        +- (0, 9.389)
    (htbot,1458.766)         +- (0, 12.135)
    (miuref,116.448)         +- (0, 56.076)
    (neris,667.949)          +- (0, 8.712)
    (nsis-ay,211.600)        +- (0, 2.951)
    (shifu,457.297)          +- (0, 2.961)
    (tinba,128.930)          +- (0, 2.310)
    (virut,767.049)          +- (0, 4.144)
    (zeus,470.147)           +- (0, 6.035)
};

\addplot+[
    red!90!black,fill=red!60!white,
    bar shift=2.1pt,
    error bars/.cd,
    y dir=both,
    y explicit relative,
    error mark options={
      rotate=90,
      mark size=0.9pt,
      line width=0.5pt
    }
] coordinates{
    (cridex,253631.113)      +- (0, 1.033)
    (geodo,2219.844)         +- (0, 27.902)
    (htbot,1292.353)         +- (0, 12.742)
    (miuref,5.519)           +- (0, 3.947)
    (neris,284.284)          +- (0, 10.365)
    (nsis-ay,185.050)        +- (0, 3.133)
    (shifu,159.062)          +- (0, 5.099)
    (tinba,98.412)           +- (0, 2.670)
    (virut,750.441)          +- (0, 4.143)
    (zeus,453.288)           +- (0, 6.093)
};

\end{axis}

\end{tikzpicture}

%% file: assets/tikz/ustc/ustc_avg_iat_ms.tikz
\begin{tikzpicture}
 
\begin{axis} [ybar=0.1pt,
height = 4cm, width = 12.5cm,
bar width = 4pt,
title=\textcolor{blue}{(\ref{app:fig:ustc_avg_iat_ms})} AVG IAT (ms) x Session,every axis title/.style={below right,at={(0,1)},draw=gray,fill=black!5,font=\fontsize{6}{6}\selectfont},
ymin = 1,
ymode=log,
label style={font=\small},
log basis y={10},
ytick distance=10,
ymajorgrids=true,
xticklabel shift={-5pt},
yticklabel shift={-2pt},
enlarge x limits=0.025,
symbolic x coords={
torrent,ftp,mysql,weibo,smb,gmail,wow,outlook,skype,facetime,geodo,neris,virut,cridex,miuref,zeus,shifu,htbot,nsis-ay,tinba
},
xtick={
torrent,ftp,mysql,weibo,smb,gmail,wow,outlook,skype,facetime,geodo,neris,virut,cridex,miuref,zeus,shifu,htbot,nsis-ay,tinba
},
x tick label style={
font=\fontsize{3.15}{1}\selectfont
},
y tick label style={font=\tiny},
xtick style={draw=none},
ytick style={draw=none},
]

\addplot+[
    blue!90!black,fill=blue!35!white,
    bar shift=-2.1pt,
    error bars/.cd,
    y dir=both,
    y explicit relative,
    error mark options={
      rotate=90,
      mark size=0.9pt,
      line width=0.5pt
    }
] coordinates {
    (facetime,0.000)         +- (0, 0.000)
    (torrent,0.00001)          +- (0, 0.000)
    (ftp,15.282)             +- (0, 1.404)
    (gmail,0.351)            +- (0, 3.849)
    (mysql,0.025)            +- (0, 5.034)
    (outlook,0.001)          +- (0, 30.684)
    (skype,0.000)            +- (0, 0.000)
    (smb,0.108)              +- (0, 27.099)
    (weibo,0.021)            +- (0, 8.076)
    (wow,0.127)              +- (0, 0.145)
};
\addplot+[
    blue!90!black,fill=blue!60!white,
    bar shift=2.1pt,
    error bars/.cd,
    y dir=both,
    y explicit relative,
    error mark options={
      rotate=90,
      mark size=0.9pt,
      line width=0.5pt
    }
] coordinates {
    (facetime,0.000)         +- (0, 0.000)
    (torrent,0.000)          +- (0, 0.000)
    (ftp,21.470)             +- (0, 1.229)
    (gmail,0.443)            +- (0, 4.348)
    (mysql,0.016)            +- (0, 7.831)
    (outlook,0.000)          +- (0, 0.000)
    (skype,0.000)            +- (0, 0.000)
    (smb,1.837)              +- (0, 7.399)
    (weibo,0.416)            +- (0, 11.263)
    (wow,0.127)              +- (0, 0.173)
};

\addplot+[
    red!90!black,fill=red!35!white,
    bar shift=-2.1pt,
    error bars/.cd,
    y dir=both,
    y explicit relative,
    error mark options={
      rotate=90,
      mark size=0.9pt,
      line width=0.5pt
    }
] coordinates{
    (cridex,31764190.000)    +- (0, 0.440)
    (geodo,6106296.000)      +- (0, 7.789)
    (htbot,201708.700)       +- (0, 32.088)
    (miuref,30162.160)       +- (0, 36.269)
    (neris,73400.730)        +- (0, 11.267)
    (nsis-ay,17745.790)      +- (0, 6.666)
    (shifu,13340.040)        +- (0, 13.323)
    (tinba,164088.500)       +- (0, 1.818)
    (virut,92997.100)        +- (0, 4.567)
    (zeus,113419.500)        +- (0, 6.877)
};

\addplot+[
    red!90!black,fill=red!60!white,
    bar shift=2.1pt,
    error bars/.cd,
    y dir=both,
    y explicit relative,
    error mark options={
      rotate=90,
      mark size=0.9pt,
      line width=0.5pt
    }
] coordinates{
    (cridex,24202960.000)    +- (0, 0.048)
    (geodo,3569236.000)      +- (0, 10.124)
    (htbot,185670.800)       +- (0, 33.972)
    (miuref,3793.133)        +- (0, 1.419)
    (neris,79802.700)        +- (0, 6.870)
    (nsis-ay,6147.955)       +- (0, 11.483)
    (shifu,323018.200)       +- (0, 3.409)
    (tinba,167654.000)       +- (0, 1.806)
    (virut,247713.000)       +- (0, 3.187)
    (zeus,368307.800)        +- (0, 6.679)
};

\end{axis}

\end{tikzpicture}

%% file: assets/tikz/ustc/ustc_protocol_dist.tikz
\begin{tikzpicture}
 
\begin{axis} [ybar=0.3pt,
height = 4cm, width = 12.5cm,
bar width = 2pt,
title=\textcolor{blue}{(\ref{app:fig:ustc_proto})} Protocol Distribution,every axis title/.style={below left,at={(1,1)},draw=gray,fill=black!5,font=\fontsize{6}{6}\selectfont},
label style={font=\small},
ymin = 1,
ymode=log,
log basis y={10},
ytick distance=10,
ymajorgrids=true,
enlarge x limits=0.025,
xticklabel shift={-5pt},
yticklabel shift={-2pt},
symbolic x coords={
torrent,ftp,mysql,weibo,smb,gmail,wow,outlook,skype,facetime,geodo,neris,virut,cridex,miuref,zeus,shifu,htbot,nsis-ay,tinba
},
xtick={
ftp,mysql,geodo,weibo,smb,neris,virut,cridex,miuref,zeus,shifu,gmail,tinba,wow,outlook,torrent,htbot,skype,nsis-ay,facetime
},
x tick label style={
font=\fontsize{3.15}{1}\selectfont,
},
y tick label style={font=\tiny},
xtick style={draw=none},
ytick style={draw=none},
]

\addplot+[
    gray!90!black,fill=gray!60!white,
] coordinates {
    (geodo,2.000)
    (htbot,2.000)
    (miuref,3.000)
    (neris,98.000)
    (nsis-ay,93.000)
    (shifu,3.000)
    (tinba,2.000)
    (virut,43.000)        
};
\addplot+[
    red!50!gray,fill=red!35!white,
] coordinates {
    (neris,1.000)
    (nsis-ay,3.000)
    (virut,1.000)
    (zeus,2.000)
};

\addplot+[
    green!90!black,fill=green!35!white,
] coordinates {
    (cridex,16378.000)
    (ftp,94886.000)
    (geodo,40278.000)
    (gmail,8629.000)
    (htbot,4689.000)
    (miuref,13426.000)
    (mysql,79141.000)
    (neris,31004.000)
    (nsis-ay,370.000)
    (outlook,7524.000)
    (shifu,409.000)
    (skype,6321.000)
    (smb,38937.000)
    (torrent,7517.000)
    (virut,30321.000)
    (weibo,39950.000)
    (wow,7883.000)
    (zeus,5780.000) 
};

\addplot+[
    blue!90!black,fill=blue!35!white,
] coordinates {
    (cridex,8.000)
    (facetime,6000.000)
    (ftp,6151.000)
    (geodo,669.000)
    (htbot,1678.000)
    (miuref,55.000)
    (mysql,6948.000)
    (neris,2787.000)
    (nsis-ay,5699.000)
    (shifu,9225.000)
    (tinba,8504.000)
    (virut,2782.000)
    (zeus,5190.000)
};

\addplot+[
    yellow!60!black,fill=yellow!35!white,
] coordinates {
    (cridex,3.000)
    (geodo,3.000)
    (htbot,3.000)
    (miuref,3.000)
    (shifu,3.000)
    (tinba,3.000)
};

\end{axis}

\end{tikzpicture}

%% file: assets/tikz/ustc/ustc_unopen_tcp_dist.tikz
\begin{tikzpicture}
 
\begin{axis} [ybar=0.3pt,
height = 4cm, width = 12.5cm,
bar width = 6pt,
title=\textcolor{blue}{(\ref{app:fig:ustc_unopen})} Unopened TCP Sessions,every axis title/.style={below right,at={(0,1)},draw=gray,fill=black!5,font=\fontsize{6}{6}\selectfont},
label style={font=\small},
ymin = 1,
ymode=log,
log basis y={10},
ytick distance=10,
ymajorgrids=true,
enlarge x limits=0.025,
xticklabel shift={-5pt},
yticklabel shift={-2pt},
symbolic x coords={
torrent,ftp,mysql,weibo,smb,gmail,wow,outlook,skype,facetime,geodo,neris,virut,cridex,miuref,zeus,shifu,htbot,nsis-ay,tinba
},
xtick={
torrent,ftp,mysql,weibo,smb,gmail,wow,outlook,skype,facetime,geodo,neris,virut,cridex,miuref,zeus,shifu,htbot,nsis-ay,tinba
},
x tick label style={
font=\fontsize{3.15}{1}\selectfont,
},
y tick label style={font=\tiny},
xtick style={draw=none},
ytick style={draw=none},
]

\addplot+[
    green!60!black,fill=green!35!white,
] coordinates {
    (torrent,0.0001) 
    (cridex,8189.000)
    (ftp,4.000)
    (geodo,34222.000)
    (gmail,22.000)
    (htbot,13.000)
    (miuref,6198.000)
    (mysql,12.000)
    (neris,22787.000)
    (nsis-ay,4.000)
    (shifu,33.000)
    (smb,1.000)
    (virut,7510.000)
    (zeus,147.000)        
    (tinba,0.0001) 
};

\end{axis}

\end{tikzpicture}

%% file: assets/tikz/boa/boa_avg_packet_size.tikz

\begin{tikzpicture}
\definecolor{cpurple}{RGB}{255,0,255}
 
\begin{axis} [ybar=0.1pt, 
height = 4cm, width = 14cm,
bar width = 2pt,
ymin = 1,
ymode=log,
title=\textcolor{blue}{(\ref{app:fig:boa_avg_pkt_size})} AVG Pkt Size x Session,every axis title/.style={below right,at={(0,1)},draw=gray,fill=black!5,font=\fontsize{4}{6}\selectfont},
log basis y={10},
ytick distance=10,
ymajorgrids=true,
enlarge x limits=0.035,
xticklabel shift={-5pt},
yticklabel shift={-2pt},
symbolic x coords={
c-fb, ff-fb, c-google, ff-google, s-google, ie-google, c-sndcld,  ff-sndcld, c-tweet, ff-tweet, s-tweet, ie-tweet, c-ytube,  ff-ytube, s-ytube, dropbox, msft, vine, tmview,
},
xtick={
dropbox, c-fb, c-google, msft, c-sndcld, tmview, c-tweet, vine, c-ytube, ff-fb, ff-google, ff-sndcld, ff-tweet, ff-ytube, ie-google, ie-tweet, s-google, s-tweet, s-ytube
},
x tick label style={
font=\fontsize{4}{1}\selectfont, rotate=45
},
y tick label style={font=\tiny},
ytick style={draw=none},
]

\addplot+[
    red!40!black,fill=red!20!orange,
    error bars/.cd,
    y dir=both,
    y explicit relative,
    error mark options={
      rotate=90,
      mark size=0.9pt,
      line width=0.5pt
    }
] coordinates {
    (c-fb,141.526)           +- (0, 0.257)
    (c-google,142.545)       +- (0, 0.418)
    (c-sndcld,95.998)        +- (0, 0.797)
    (c-tweet,214.701)        +- (0, 0.513)
    (c-ytube,76.535)         +- (0, 0.232)
    (ff-fb,225.227)          +- (0, 0.563)
    (ff-google,211.836)      +- (0, 0.692)
    (ff-sndcld,103.221)      +- (0, 0.591)
    (ff-tweet,246.140)       +- (0, 0.499)
    (ff-ytube,68.910)        +- (0, 0.116)
    (msft,103.453)           +- (0, 0.185)
};
\addplot+[
    red!40!black,fill=red!60!orange,
    error bars/.cd,
    y dir=both,
    y explicit relative,
    error mark options={
      rotate=90,
      mark size=0.9pt,
      line width=0.5pt
    }
] coordinates {
    (c-fb,1089.614)          +- (0, 0.342)
    (c-google,866.274)       +- (0, 0.507)
    (c-sndcld,1705.637)      +- (0, 0.158)
    (c-tweet,983.753)        +- (0, 0.731)
    (c-ytube,1678.010)       +- (0, 0.117)
    (ff-fb,1233.772)         +- (0, 0.388)
    (ff-google,327.608)      +- (0, 1.002)
    (ff-sndcld,1623.845)     +- (0, 0.286)
    (ff-tweet,766.729)       +- (0, 0.832)
    (ff-ytube,1783.290)      +- (0, 0.047)
    (msft,1025.170)          +- (0, 0.194)
    (vine,1642.573)          +- (0, 0.232)
};

\addplot+[
    purple!20!black,fill=cpurple!50!pink,
    bar shift=0.0pt,
    error bars/.cd,
    y dir=both,
    y explicit relative,
    error mark options={
      rotate=90,
      mark size=0.9pt,
      line width=0.5pt
    }
] coordinates {
    (c-google,110.485)       +- (0, 0.079)
    (c-tweet,117.089)        +- (0, 0.426)
    (s-google,123.154)       +- (0, 0.440)
    (s-tweet,420.320)        +- (0, 0.409)
    (s-ytube,88.941)         +- (0, 0.538)
};
\addplot+[
    purple!20!black,fill=cpurple!80!black,
    bar shift=2pt,
    error bars/.cd,
    y dir=both,
    y explicit relative,
    error mark options={
      rotate=90,
      mark size=0.9pt,
      line width=0.5pt
    }
] coordinates {
    (c-google,352.382)       +- (0, 0.194)
    (c-tweet,1077.225)       +- (0, 0.303)
    (s-google,998.911)       +- (0, 0.348)
    (s-tweet,518.594)        +- (0, 0.832)
    (s-ytube,1382.562)       +- (0, 0.113)
};

\addplot+[
    blue!90!black,fill=blue!35!white,
    bar shift=5.1pt,
    error bars/.cd,
    y dir=both,
    y explicit relative,
    error mark options={
      rotate=90,
      mark size=0.9pt,
      line width=0.5pt
    }
] coordinates {
    (c-google,132.790)       +- (0, 0.449)
    (c-tweet,814.305)        +- (0, 0.436)
    (ff-google,157.399)      +- (0, 1.106)
    (ff-tweet,307.513)       +- (0, 1.228)
    (ie-google,192.493)      +- (0, 0.146)
    (ie-tweet,412.359)       +- (0, 0.507)
    (dropbox,68.829)         +- (0, 1.302)
    (msft,451.288)           +- (0, 0.960)
    (tmview,142.833)         +- (0, 0.045)
    (vine,85.453)            +- (0, 0.640)
};
\addplot+[
    blue!90!black,fill=blue!60!white,
     bar shift=7.1pt,
    error bars/.cd,
    y dir=both,
    y explicit relative,
    error mark options={
      rotate=90,
      mark size=0.9pt,
      line width=0.5pt
    }
] coordinates {
    (c-google,301.014)       +- (0, 0.396)
    (c-tweet,441.439)        +- (0, 1.203)
    (ff-google,174.925)      +- (0, 0.633)
    (ff-tweet,757.856)       +- (0, 0.666)
    (ie-google,520.885)      +- (0, 0.266)
    (ie-tweet,829.182)       +- (0, 0.553)
    (dropbox,1396.967)       +- (0, 0.092)
    (msft,657.687)           +- (0, 0.517)
    (tmview,781.744)         +- (0, 0.149)
    (vine,1337.916)          +- (0, 0.169)
};

\end{axis}

\end{tikzpicture}

%% file: assets/tikz/boa/boa_avg_packets.tikz

\begin{tikzpicture}
\definecolor{cpurple}{RGB}{255,0,255}
 
\begin{axis} [ybar=0.1pt, 
height = 4cm, width = 14cm,
bar width = 2pt,
ymin = 1,
ymode=log,
title=\textcolor{blue}{(\ref{app:fig:boa_avg_pkts})} AVG Packets x Session,every axis title/.style={below right,at={(0,1)},draw=gray,fill=black!5,font=\fontsize{6}{6}\selectfont},
log basis y={10},
ytick distance=10,
ymajorgrids=true,
enlarge x limits=0.035,
xticklabel shift={-5pt},
yticklabel shift={-2pt},
symbolic x coords={
c-fb, ff-fb, c-google, ff-google, s-google, ie-google, c-sndcld,  ff-sndcld, c-tweet, ff-tweet, s-tweet, ie-tweet, c-ytube,  ff-ytube, s-ytube, dropbox, msft, vine, tmview,
},
xtick={
dropbox, c-fb, c-google, msft, c-sndcld, tmview, c-tweet, vine, c-ytube, ff-fb, ff-google, ff-sndcld, ff-tweet, ff-ytube, ie-google, ie-tweet, s-google, s-tweet, s-ytube
},
x tick label style={
font=\fontsize{4}{1}\selectfont, rotate=45
},
y tick label style={font=\tiny},
ytick style={draw=none},
]

\addplot+[
    red!40!black,fill=red!20!orange,
    error bars/.cd,
    y dir=both,
    y explicit relative,
    error mark options={
      rotate=90,
      mark size=0.9pt,
      line width=0.5pt
    }
] coordinates {
    (c-fb,26.156)            +- (0, 1.892)
    (c-google,80.069)        +- (0, 3.105)
    (c-sndcld,168.316)       +- (0, 2.687)
    (c-tweet,2346.678)       +- (0, 3.413)
    (c-ytube,2283.424)       +- (0, 1.576)
    (ff-fb,78.181)           +- (0, 1.264)
    (ff-google,31.571)       +- (0, 2.473)
    (ff-sndcld,144.591)      +- (0, 2.857)
    (ff-tweet,1166.060)      +- (0, 4.780)
    (ff-ytube,5648.234)      +- (0, 4.771)
    (msft,11.036)            +- (0, 0.217)
    (vine,112.867)           +- (0, 2.562)

};
\addplot+[
    red!40!black,fill=red!60!orange,
    error bars/.cd,
    y dir=both,
    y explicit relative,
    error mark options={
      rotate=90,
      mark size=0.9pt,
      line width=0.5pt
    }
] coordinates {
    (c-fb,24.646)            +- (0, 2.308)
    (c-google,110.442)       +- (0, 3.364)
    (c-sndcld,293.171)       +- (0, 3.656)
    (c-tweet,2798.138)       +- (0, 3.224)
    (c-ytube,3688.353)       +- (0, 1.636)
    (ff-fb,83.133)           +- (0, 1.500)
    (ff-google,36.267)       +- (0, 2.743)
    (ff-sndcld,183.413)      +- (0, 3.440)
    (ff-tweet,1196.758)      +- (0, 4.617)
    (ff-ytube,10750.253)     +- (0, 4.844)
    (msft,8.179)             +- (0, 0.354)
    (vine,127.378)           +- (0, 2.607)
};

\addplot+[
    purple!20!black,fill=cpurple!50!pink,
    bar shift=0.0pt,
    error bars/.cd,
    y dir=both,
    y explicit relative,
    error mark options={
      rotate=90,
      mark size=0.9pt,
      line width=0.5pt
    }
] coordinates {
    (c-google,27.368)        +- (0, 0.286)
    (c-tweet,147.772)        +- (0, 1.266)
    (s-google,61.072)        +- (0, 2.486)
    (s-tweet,232.960)        +- (0, 9.106)
    (s-ytube,699.496)        +- (0, 2.477)
};
\addplot+[
    purple!20!black,fill=cpurple!80!black,
    bar shift=2pt,
    error bars/.cd,
    y dir=both,
    y explicit relative,
    error mark options={
      rotate=90,
      mark size=0.9pt,
      line width=0.5pt
    }
] coordinates {
    (c-google,26.053)        +- (0, 0.358)
    (c-tweet,187.281)        +- (0, 1.197)
    (s-google,75.474)        +- (0, 3.910)
    (s-tweet,248.297)        +- (0, 8.606)
    (s-ytube,1748.191)       +- (0, 3.070)
};

\addplot+[
    blue!90!black,fill=blue!35!white,
    bar shift=5.1pt,
    error bars/.cd,
    y dir=both,
    y explicit relative,
    error mark options={
      rotate=90,
      mark size=0.9pt,
      line width=0.5pt
    }
] coordinates {
    (c-google,22.945)        +- (0, 0.629)
    (c-tweet,3382.610)       +- (0, 7.845)
    (ff-google,28.880)       +- (0, 1.269)
    (ff-tweet,573.320)       +- (0, 4.984)
    (ie-google,17.593)       +- (0, 0.370)
    (ie-tweet,150.234)       +- (0, 5.636)
    (dropbox,1354.079)       +- (0, 5.717)
    (msft,16.486)            +- (0, 3.173)
    (tmview,10.571)          +- (0, 0.073)
};
\addplot+[
    blue!90!black,fill=blue!60!white,
     bar shift=7.1pt,
    error bars/.cd,
    y dir=both,
    y explicit relative,
    error mark options={
      rotate=90,
      mark size=0.9pt,
      line width=0.5pt
    }
] coordinates {
     (c-google,26.778)        +- (0, 0.739)
    (c-tweet,2767.666)       +- (0, 5.538)
    (ff-google,34.869)       +- (0, 1.278)
    (ff-tweet,1001.995)      +- (0, 4.571)
    (ie-google,20.667)       +- (0, 0.386)
    (ie-tweet,207.553)       +- (0, 8.211)
    (dropbox,2475.921)       +- (0, 5.746)
    (msft,19.615)            +- (0, 3.688)
    (tmview,11.918)          +- (0, 0.124)
    (vine,192.899)           +- (0, 2.394)
};

\end{axis}

\end{tikzpicture}

%% file: assets/tikz/boa/boa_avg_bytes.tikz

\begin{tikzpicture}
\definecolor{cpurple}{RGB}{255,0,255}
 
\begin{axis} [ybar=0.1pt, 
height = 4cm, width = 14cm,
bar width = 2pt,
ymin = 1,
ymode=log,
title=\textcolor{blue}{(\ref{app:fig:boa_avg_bytes})} AVG Bytes x Session,every axis title/.style={below right,at={(0,1)},draw=gray,fill=black!5,font=\fontsize{6}{6}\selectfont},
log basis y={10},
ytick distance=10,
ymajorgrids=true,
enlarge x limits=0.035,
xticklabel shift={-5pt},
yticklabel shift={-2pt},
symbolic x coords={
c-fb, ff-fb, c-google, ff-google, s-google, ie-google, c-sndcld,  ff-sndcld, c-tweet, ff-tweet, s-tweet, ie-tweet, c-ytube,  ff-ytube, s-ytube, dropbox, msft, vine, tmview,
},
xtick={
dropbox, c-fb, c-google, msft, c-sndcld, tmview, c-tweet, vine, c-ytube, ff-fb, ff-google, ff-sndcld, ff-tweet, ff-ytube, ie-google, ie-tweet, s-google, s-tweet, s-ytube
},
x tick label style={
font=\fontsize{4}{1}\selectfont, rotate=45
},
y tick label style={font=\tiny},
ytick style={draw=none},
]

\addplot+[
    red!40!black,fill=red!20!orange,
    error bars/.cd,
    y dir=both,
    y explicit relative,
    error mark options={
      rotate=90,
      mark size=0.9pt,
      line width=0.5pt
    }
] coordinates {
    (c-fb,3701.781)          +- (0, 2.128)
    (c-google,11413.404)     +- (0, 3.572)
    (c-sndcld,16158.032)     +- (0, 1.901)
    (c-tweet,503834.700)     +- (0, 4.363)
    (c-ytube,174760.856)     +- (0, 1.496)
    (ff-fb,17608.446)        +- (0, 1.580)
    (ff-google,6687.952)     +- (0, 2.047)
    (ff-sndcld,14924.809)    +- (0, 1.870)
    (ff-tweet,287014.647)    +- (0, 6.066)
    (ff-ytube,389217.610)    +- (0, 4.618)
    (msft,1141.679)          +- (0, 0.406)
    (vine,10715.022)         +- (0, 2.011)

};
\addplot+[
    red!40!black,fill=red!60!orange,
    error bars/.cd,
    y dir=both,
    y explicit relative,
    error mark options={
      rotate=90,
      mark size=0.9pt,
      line width=0.5pt
    }
] coordinates {
    (c-fb,26854.440)         +- (0, 2.848)
    (c-google,95672.760)     +- (0, 4.109)
    (c-sndcld,500043.400)    +- (0, 3.770)
    (c-tweet,2752677.000)    +- (0, 3.618)
    (c-ytube,6189091.000)    +- (0, 1.661)
    (ff-fb,102566.600)       +- (0, 1.990)
    (ff-google,11881.370)    +- (0, 8.198)
    (ff-sndcld,297834.800)   +- (0, 3.960)
    (ff-tweet,917589.000)    +- (0, 3.394)
    (ff-ytube,19170820.000)  +- (0, 4.926)
    (msft,8384.429)          +- (0, 0.110)
    (vine,209227.200)        +- (0, 2.836)
};

\addplot+[
    purple!20!black,fill=cpurple!50!pink,
    bar shift=0.0pt,
    error bars/.cd,
    y dir=both,
    y explicit relative,
    error mark options={
      rotate=90,
      mark size=0.9pt,
      line width=0.5pt
    }
] coordinates {
    (c-google,3023.789)      +- (0, 0.262)
    (c-tweet,17302.404)      +- (0, 1.924)
    (s-google,7521.310)      +- (0, 1.639)
    (s-tweet,97917.683)      +- (0, 10.966)
    (s-ytube,62213.634)      +- (0, 1.990)

};
\addplot+[
    purple!20!black,fill=cpurple!80!black,
    bar shift=2pt,
    error bars/.cd,
    y dir=both,
    y explicit relative,
    error mark options={
      rotate=90,
      mark size=0.9pt,
      line width=0.5pt
    }
] coordinates {
    (c-google,9180.474)      +- (0, 0.516)
    (c-tweet,201743.400)     +- (0, 1.107)
    (s-google,75391.830)     +- (0, 5.327)
    (s-tweet,128765.500)     +- (0, 4.227)
    (s-ytube,2416982.000)    +- (0, 3.140)
};

\addplot+[
    blue!90!black,fill=blue!35!white,
    bar shift=5.1pt,
    error bars/.cd,
    y dir=both,
    y explicit relative,
    error mark options={
      rotate=90,
      mark size=0.9pt,
      line width=0.5pt
    }
] coordinates {
    (c-google,3046.863)      +- (0, 0.661)
    (c-tweet,2754477.000)    +- (0, 9.445)
    (ff-google,4545.670)     +- (0, 1.284)
    (ff-tweet,176303.700)    +- (0, 8.277)
    (ie-google,3386.444)     +- (0, 0.441)
    (ie-tweet,61950.260)     +- (0, 0.869)
    (dropbox,93199.990)      +- (0, 4.886)
    (msft,7439.818)          +- (0, 7.838)
    (tmview,1509.946)        +- (0, 0.028)
    (vine,7075.261)          +- (0, 1.674)

};
\addplot+[
    blue!90!black,fill=blue!60!white,
     bar shift=7.1pt,
    error bars/.cd,
    y dir=both,
    y explicit relative,
    error mark options={
      rotate=90,
      mark size=0.9pt,
      line width=0.5pt
    }
] coordinates {
    (c-google,8060.668)      +- (0, 0.646)
    (c-tweet,1221756.000)    +- (0, 4.579)
    (ff-google,6099.416)     +- (0, 0.979)
    (ff-tweet,759368.000)    +- (0, 5.693)
    (ie-google,10764.960)    +- (0, 0.557)
    (ie-tweet,172099.600)    +- (0, 13.979)
    (dropbox,3458780.000)    +- (0, 5.809)
    (msft,12900.370)         +- (0, 5.334)
    (tmview,9317.109)        +- (0, 0.263)
    (vine,258082.000)        +- (0, 2.524)
};

\end{axis}

\end{tikzpicture}

%% file: assets/tikz/boa/boa_avg_duration_s.tikz

\begin{tikzpicture}
\definecolor{cpurple}{RGB}{255,0,255}
 
\begin{axis} [ybar=0.1pt, 
height = 4cm, width = 14cm,
bar width = 2pt,
ymin = 1,
ymode=log,
title=\textcolor{blue}{(\ref{app:fig:boa_avg_dur_s})} AVG Duration (sec) x Session,every axis title/.style={below right,at={(0,1)},draw=gray,fill=black!5,font=\fontsize{6}{6}\selectfont},
log basis y={10},
ytick distance=10,
ymajorgrids=true,
enlarge x limits=0.035,
xticklabel shift={-5pt},
yticklabel shift={-2pt},
symbolic x coords={
c-fb, ff-fb, c-google, ff-google, s-google, ie-google, c-sndcld,  ff-sndcld, c-tweet, ff-tweet, s-tweet, ie-tweet, c-ytube,  ff-ytube, s-ytube, dropbox, msft, vine, tmview,
},
xtick={
dropbox, c-fb, c-google, msft, c-sndcld, tmview, c-tweet, vine, c-ytube, ff-fb, ff-google, ff-sndcld, ff-tweet, ff-ytube, ie-google, ie-tweet, s-google, s-tweet, s-ytube
},
x tick label style={
font=\fontsize{4}{1}\selectfont, rotate=45
},
y tick label style={font=\tiny},
ytick style={draw=none},
]

\addplot+[
    red!40!black,fill=red!20!orange,
    error bars/.cd,
    y dir=both,
    y explicit relative,
    error mark options={
      rotate=90,
      mark size=0.9pt,
      line width=0.5pt
    }
] coordinates {
    (c-fb,29.520)            +- (0, 1.153)
    (c-google,279.694)       +- (0, 1.078)
    (c-sndcld,159.206)       +- (0, 0.944)
    (c-tweet,3485.675)       +- (0, 4.021)
    (c-ytube,149.741)        +- (0, 2.670)
    (ff-fb,123.458)          +- (0, 1.403)
    (ff-google,260.573)      +- (0, 0.478)
    (ff-sndcld,119.174)      +- (0, 0.521)
    (ff-tweet,1963.929)      +- (0, 5.461)
    (ff-ytube,33.632)        +- (0, 3.889)
    (msft,26.529)            +- (0, 1.673)
    (vine,53.253)            +- (0, 1.176)

};
\addplot+[
    red!40!black,fill=red!60!orange,
    error bars/.cd,
    y dir=both,
    y explicit relative,
    error mark options={
      rotate=90,
      mark size=0.9pt,
      line width=0.5pt
    }
] coordinates {
    (c-fb,20.701)            +- (0, 1.844)
    (c-google,277.821)       +- (0, 1.091)
    (c-sndcld,157.806)       +- (0, 0.961)
    (c-tweet,3486.963)       +- (0, 4.019)
    (c-ytube,149.693)        +- (0, 2.671)
    (ff-fb,123.434)          +- (0, 1.403)
    (ff-google,260.493)      +- (0, 0.479)
    (ff-sndcld,119.103)      +- (0, 0.522)
    (ff-tweet,1963.783)      +- (0, 5.461)
    (ff-ytube,30.817)        +- (0, 4.045)
    (msft,13.103)            +- (0, 3.628)
    (vine,52.794)            +- (0, 1.192)
};

\addplot+[
    purple!20!black,fill=cpurple!50!pink,
    bar shift=0.0pt,
    error bars/.cd,
    y dir=both,
    y explicit relative,
    error mark options={
      rotate=90,
      mark size=0.9pt,
      line width=0.5pt
    }
] coordinates {
    (c-google,152.490)       +- (0, 0.783)
    (c-tweet,200.991)        +- (0, 1.639)
    (s-google,73.827)        +- (0, 1.478)
    (s-tweet,486.530)        +- (0, 7.642)
    (s-ytube,102.839)        +- (0, 1.075)
};
\addplot+[
    purple!20!black,fill=cpurple!80!black,
    bar shift=2pt,
    error bars/.cd,
    y dir=both,
    y explicit relative,
    error mark options={
      rotate=90,
      mark size=0.9pt,
      line width=0.5pt
    }
] coordinates {
    (c-google,152.498)       +- (0, 0.783)
    (c-tweet,200.856)        +- (0, 1.639)
    (s-google,74.071)        +- (0, 1.474)
    (s-tweet,484.158)        +- (0, 7.680)
    (s-ytube,103.393)        +- (0, 1.070)
};

\addplot+[
    blue!90!black,fill=blue!35!white,
    bar shift=5.1pt,
    error bars/.cd,
    y dir=both,
    y explicit relative,
    error mark options={
      rotate=90,
      mark size=0.9pt,
      line width=0.5pt
    }
] coordinates {
    (c-google,296.577)       +- (0, 1.183)
    (c-tweet,1550.565)       +- (0, 3.931)
    (ff-google,393.234)      +- (0, 1.761)
    (ff-tweet,1127.871)      +- (0, 5.033)
    (ie-google,72.903)       +- (0, 0.638)
    (ie-tweet,424.667)       +- (0, 4.080)
    (dropbox,742.032)        +- (0, 1.837)
    (msft,129.460)           +- (0, 15.640)
    (tmview,0.557)           +- (0, 0.601)
    (vine,77.758)            +- (0, 0.861)
};
\addplot+[
    blue!90!black,fill=blue!60!white,
     bar shift=7.1pt,
    error bars/.cd,
    y dir=both,
    y explicit relative,
    error mark options={
      rotate=90,
      mark size=0.9pt,
      line width=0.5pt
    }
] coordinates {
    (c-google,224.451)       +- (0, 1.137)
    (c-tweet,1550.168)       +- (0, 3.932)
    (ff-google,387.962)      +- (0, 1.781)
    (ff-tweet,1128.547)      +- (0, 5.030)
    (ie-google,72.841)       +- (0, 0.639)
    (ie-tweet,418.270)       +- (0, 4.146)
    (dropbox,464.880)        +- (0, 2.249)
    (msft,124.294)           +- (0, 15.780)
    (tmview,0.467)           +- (0, 0.705)
    (vine,51.377)            +- (0, 1.056)
};

\end{axis}

\end{tikzpicture}

%% file: assets/tikz/boa/boa_avg_iat_ms.tikz

\begin{tikzpicture}
\definecolor{cpurple}{RGB}{255,0,255}
 
\begin{axis} [ybar=0.1pt, 
height = 4cm, width = 14cm,
bar width = 2pt,
ymin = 1,
ymode=log,
title=\textcolor{blue}{(\ref{app:fig:boa_avg_iat_ms})} AVG IAT (ms) x Session,every axis title/.style={below right,at={(0,1)},draw=gray,fill=black!5,font=\fontsize{6}{6}\selectfont},
log basis y={10},
ytick distance=10,
ymajorgrids=true,
enlarge x limits=0.035,
xticklabel shift={-5pt},
yticklabel shift={-2pt},
symbolic x coords={
c-fb, ff-fb, c-google, ff-google, s-google, ie-google, c-sndcld,  ff-sndcld, c-tweet, ff-tweet, s-tweet, ie-tweet, c-ytube,  ff-ytube, s-ytube, dropbox, msft, vine, tmview,
},
xtick={
dropbox, c-fb, c-google, msft, c-sndcld, tmview, c-tweet, vine, c-ytube, ff-fb, ff-google, ff-sndcld, ff-tweet, ff-ytube, ie-google, ie-tweet, s-google, s-tweet, s-ytube
},
x tick label style={
font=\fontsize{4}{1}\selectfont, rotate=45
},
y tick label style={font=\tiny},
ytick style={draw=none},
]

\addplot+[
    red!40!black,fill=red!20!orange,
    error bars/.cd,
    y dir=both,
    y explicit relative,
    error mark options={
      rotate=90,
      mark size=0.9pt,
      line width=0.5pt
    }
] coordinates {
    (c-fb,1196.905)          +- (0, 0.745)
    (c-google,3598.679)      +- (0, 1.184)
    (c-sndcld,975.633)       +- (0, 2.341)
    (c-tweet,1487.578)       +- (0, 0.803)
    (c-ytube,66.145)         +- (0, 7.610)
    (ff-fb,1605.429)         +- (0, 0.996)
    (ff-google,8626.618)     +- (0, 0.606)
    (ff-sndcld,843.207)      +- (0, 1.650)
    (ff-tweet,1685.903)      +- (0, 0.392)
    (ff-ytube,5.981)         +- (0, 16.636)
    (msft,2601.201)          +- (0, 1.119)
    (vine,497.742)           +- (0, 3.567)
};
\addplot+[
    red!40!black,fill=red!60!orange,
    error bars/.cd,
    y dir=both,
    y explicit relative,
    error mark options={
      rotate=90,
      mark size=0.9pt,
      line width=0.5pt
    }
] coordinates {
    (c-fb,863.720)           +- (0, 0.996)
    (c-google,2579.896)      +- (0, 1.364)
    (c-sndcld,558.706)       +- (0, 3.788)
    (c-tweet,1247.987)       +- (0, 1.008)
    (c-ytube,40.871)         +- (0, 8.640)
    (ff-fb,1511.652)         +- (0, 1.188)
    (ff-google,7471.123)     +- (0, 0.637)
    (ff-sndcld,668.288)      +- (0, 2.232)
    (ff-tweet,1642.578)      +- (0, 0.480)
    (ff-ytube,2.878)         +- (0, 21.046)
    (msft,1698.124)          +- (0, 2.287)
    (vine,440.163)           +- (0, 4.176)
};

\addplot+[
    purple!20!black,fill=cpurple!50!pink,
    bar shift=0.0pt,
    error bars/.cd,
    y dir=both,
    y explicit relative,
    error mark options={
      rotate=90,
      mark size=0.9pt,
      line width=0.5pt
    }
] coordinates {
    (c-google,5760.741)      +- (0, 0.648)
    (c-tweet,1378.880)       +- (0, 2.432)
    (s-google,1231.703)      +- (0, 1.630)
    (s-tweet,2107.611)       +- (0, 1.286)
    (s-ytube,149.130)        +- (0, 4.680)
};
\addplot+[
    purple!20!black,fill=cpurple!80!black,
    bar shift=2pt,
    error bars/.cd,
    y dir=both,
    y explicit relative,
    error mark options={
      rotate=90,
      mark size=0.9pt,
      line width=0.5pt
    }
] coordinates {
    (c-google,6061.704)      +- (0, 0.645)
    (c-tweet,1087.725)       +- (0, 2.924)
    (s-google,1000.642)      +- (0, 1.920)
    (s-tweet,1969.075)       +- (0, 1.412)
    (s-ytube,59.964)         +- (0, 7.372)
};

\addplot+[
    blue!90!black,fill=blue!35!white,
    bar shift=5.1pt,
    error bars/.cd,
    y dir=both,
    y explicit relative,
    error mark options={
      rotate=90,
      mark size=0.9pt,
      line width=0.5pt
    }
] coordinates {
    (c-google,13528.785)     +- (0, 1.036)
    (c-tweet,459.992)        +- (0, 2.422)
    (ff-google,14055.619)    +- (0, 0.357)
    (ff-tweet,1971.388)      +- (0, 0.646)
    (ie-google,4354.842)     +- (0, 0.416)
    (ie-tweet,2851.764)      +- (0, 4.117)
    (dropbox,580.643)        +- (0, 18.112)
    (msft,8102.292)          +- (0, 6.069)
    (tmview,58.240)          +- (0, 0.593)
    (vine,988.517)           +- (0, 2.538)
};
\addplot+[
    blue!90!black,fill=blue!60!white,
     bar shift=7.1pt,
    error bars/.cd,
    y dir=both,
    y explicit relative,
    error mark options={
      rotate=90,
      mark size=0.9pt,
      line width=0.5pt
    }
] coordinates {
    (c-google,8764.845)      +- (0, 0.649)
    (c-tweet,561.549)        +- (0, 1.792)
    (ff-google,11431.306)    +- (0, 0.348)
    (ff-tweet,1127.647)      +- (0, 0.718)
    (ie-google,3685.606)     +- (0, 0.500)
    (ie-tweet,2032.010)      +- (0, 5.052)
    (dropbox,191.323)        +- (0, 9.601)
    (msft,6619.927)          +- (0, 8.813)
    (tmview,42.818)          +- (0, 0.707)
    (vine,284.074)           +- (0, 4.691)
};

\end{axis}

\end{tikzpicture}

%% file: assets/tikz/iscx/iscx_avg_packet_size.tikz

\begin{tikzpicture}

\begin{axis} [ybar=0.1pt,
height = 4cm, width = 12cm,
bar width = 2pt,
ymin = 1,
ymode=log,
title=\textcolor{blue}{(\ref{app:fig:iscx_avg_pkt_size})} AVG Pkt Size x Session,every axis title/.style={below right,at={(0,1)},draw=gray,fill=black!5,font=\fontsize{6}{6}\selectfont},
log basis y={10},
ytick distance=10,
ymajorgrids=true,
enlarge x limits=0.035,
xticklabel shift={-5pt},
yticklabel shift={-2pt},
symbolic x coords={
a-fb,a-hng,a-skype,a-sptfy,a-vpbust,c-aim,c-fb, c-gmail, c-hng, c-icq, c-skype, e-email, ft-ftps, ft-scp, ft-sftp, ft-skype, p2p-torr, v-fb, v-hng, v-ntflx, v-skype, v-vimeo, v-ytube
},
xtick={
a-fb,a-hng,a-skype,a-sptfy,a-vpbust,c-aim,c-fb, c-gmail, c-hng, c-icq, c-skype, e-email, ft-ftps, ft-scp, ft-sftp, ft-skype, p2p-torr, v-fb, v-hng, v-ntflx, v-skype, v-vimeo, v-ytube
},
x tick label style={
font=\fontsize{3.5}{1}\selectfont, rotate=45, anchor=north east
},
y tick label style={font=\tiny},
xtick style={draw=none},
ytick style={draw=none},
]

\addplot+[
    blue!90!black,fill=blue!35!white,
    error bars/.cd,
    y dir=both,
    y explicit relative,
    error mark options={
      rotate=90,
      mark size=0.9pt,
      line width=0.5pt
    }
] coordinates {
    (a-fb,137.999)           +- (0, 0.365)
    (a-hng,145.466)          +- (0, 0.453)
    (a-skype,122.680)        +- (0, 0.273)
    (a-sptfy,90.188)         +- (0, 0.694)
    (a-vpbust,124.532)       +- (0, 0.099)
    (c-aim,124.927)          +- (0, 0.686)
    (c-fb,276.111)           +- (0, 0.586)
    (c-gmail,573.213)        +- (0, 0.505)
    (c-hng,363.427)          +- (0, 0.652)
    (c-icq,101.460)          +- (0, 0.595)
    (c-skype,177.190)        +- (0, 0.847)
    (e-email,102.708)        +- (0, 0.506)
    (ft-ftps,113.592)        +- (0, 2.414)
    (ft-scp,13388.555)       +- (0, 0.798)
    (ft-sftp,2984.509)       +- (0, 1.910)
    (ft-skype,717.567)       +- (0, 0.654)
    (p2p-torr,65.304)        +- (0, 0.736)
    (v-fb,917.897)           +- (0, 0.432)
    (v-hng,866.473)          +- (0, 0.216)
    (v-ntflx,2386.122)       +- (0, 0.773)
    (v-skype,650.048)        +- (0, 0.510)
    (v-vimeo,1761.510)       +- (0, 1.369)
    (v-ytube,74.277)         +- (0, 0.316)
};
\addplot+[
    blue!90!black,fill=blue!60!white,
    error bars/.cd,
    y dir=both,
    y explicit relative,
    error mark options={
      rotate=90,
      mark size=0.9pt,
      line width=0.5pt
    }
] coordinates {
    (a-fb,113.626)           +- (0, 0.384)
    (a-hng,159.152)          +- (0, 0.380)
    (a-skype,148.671)        +- (0, 0.510)
    (a-sptfy,3207.715)       +- (0, 0.270)
    (a-vpbust,126.368)       +- (0, 0.189)
    (c-aim,452.885)          +- (0, 1.056)
    (c-fb,390.054)           +- (0, 0.709)
    (c-gmail,231.090)        +- (0, 1.220)
    (c-hng,500.361)          +- (0, 0.958)
    (c-icq,159.738)          +- (0, 0.671)
    (c-skype,287.262)        +- (0, 0.609)
    (e-email,142.562)        +- (0, 1.288)
    (ft-ftps,1191.751)       +- (0, 0.478)
    (ft-scp,1760.219)        +- (0, 2.032)
    (ft-sftp,757.199)        +- (0, 1.211)
    (ft-skype,514.608)       +- (0, 0.603)
    (p2p-torr,1357.849)      +- (0, 0.149)
    (v-fb,930.489)           +- (0, 0.408)
    (v-hng,871.244)          +- (0, 0.200)
    (v-ntflx,1869.082)       +- (0, 1.070)
    (v-skype,1057.768)       +- (0, 0.408)
    (v-vimeo,4360.960)       +- (0, 0.458)
    (v-ytube,3195.070)       +- (0, 0.239)
};

\addplot+[
    green!20!black,fill=green!35!white,
    error bars/.cd,
    y dir=both,
    y explicit relative,
    error mark options={
      rotate=90,
      mark size=0.9pt,
      line width=0.5pt
    }
] coordinates {
    (a-fb,318.901)           +- (0, 0.768)
    (a-hng,152.762)          +- (0, 0.487)
    (a-skype,129.164)        +- (0, 0.070)
    (a-sptfy,65.463)         +- (0, 0.874)
    (a-vpbust,124.815)       +- (0, 0.096)
    (c-aim,137.958)          +- (0, 0.341)
    (c-fb,330.159)           +- (0, 0.457)
    (c-gmail,0.000)          +- (0, 0.000)
    (c-hng,282.294)          +- (0, 0.862)
    (c-icq,99.177)           +- (0, 0.420)
    (c-skype,129.556)        +- (0, 0.673)
    (e-email,210.246)        +- (0, 0.834)
    (ft-ftps,945.683)        +- (0, 0.632)
    (ft-scp,0.000)           +- (0, 0.000)
    (ft-sftp,1193.095)       +- (0, 0.328)
    (ft-skype,210.073)       +- (0, 0.650)
    (p2p-torr,56.994)        +- (0, 0.080)
    (v-fb,0.000)             +- (0, 0.000)
    (v-hng,0.000)            +- (0, 0.000)
    (v-ntflx,56.949)         +- (0, 0.874)
    (v-skype,0.000)          +- (0, 0.000)
    (v-vimeo,60.165)         +- (0, 0.843)
    (v-ytube,64.099)         +- (0, 0.603)
};
\addplot+[
    green!20!black,fill=green!25!gray,
    error bars/.cd,
    y dir=both,
    y explicit relative,
    error mark options={
      rotate=90,
      mark size=0.9pt,
      line width=0.5pt
    }
] coordinates {
    (a-fb,508.434)           +- (0, 0.669)
    (a-hng,163.772)          +- (0, 1.057)
    (a-skype,103.046)        +- (0, 0.311)
    (a-sptfy,1318.689)       +- (0, 0.098)
    (a-vpbust,128.586)       +- (0, 0.469)
    (c-aim,130.509)          +- (0, 0.580)
    (c-fb,367.818)           +- (0, 0.496)
    (c-gmail,0.000)          +- (0, 0.000)
    (c-hng,534.899)          +- (0, 0.882)
    (c-icq,98.809)           +- (0, 0.461)
    (c-skype,93.597)         +- (0, 0.797)
    (e-email,493.264)        +- (0, 1.027)
    (ft-ftps,871.997)        +- (0, 0.715)
    (ft-scp,0.000)           +- (0, 0.000)
    (ft-sftp,432.081)        +- (0, 1.315)
    (ft-skype,169.214)       +- (0, 0.965)
    (p2p-torr,1349.340)      +- (0, 0.039)
    (v-fb,0.000)             +- (0, 0.000)
    (v-hng,0.000)            +- (0, 0.000)
    (v-ntflx,1355.394)       +- (0, 0.043)
    (v-skype,0.000)          +- (0, 0.000)
    (v-vimeo,1354.489)       +- (0, 0.052)
    (v-ytube,1341.213)       +- (0, 0.085)
};

\end{axis}

\end{tikzpicture}

%% file: assets/tikz/iscx/iscx_avg_packets.tikz

\begin{tikzpicture}
 
\begin{axis} [ybar=0.1pt,
height = 4cm, width = 12cm,
bar width = 2pt,
ymin = 1,
ymode=log,
title=\textcolor{blue}{(\ref{app:fig:iscx_avg_pkts})} AVG Pkts x Session,every axis title/.style={below right,at={(0,1)},draw=gray,fill=black!5,font=\fontsize{6}{6}\selectfont},
log basis y={10},
ytick distance=10,
ymajorgrids=true,
enlarge x limits=0.035,
xticklabel shift={-5pt},
yticklabel shift={-2pt},
symbolic x coords={
a-fb,a-hng,a-skype,a-sptfy,a-vpbust,c-aim,c-fb, c-gmail, c-hng, c-icq, c-skype, e-email, ft-ftps, ft-scp, ft-sftp, ft-skype, p2p-torr, v-fb, v-hng, v-ntflx, v-skype, v-vimeo, v-ytube
},
xtick={
a-fb,a-hng,a-skype,a-sptfy,a-vpbust,c-aim,c-fb, c-gmail, c-hng, c-icq, c-skype, e-email, ft-ftps, ft-scp, ft-sftp, ft-skype, p2p-torr, v-fb, v-hng, v-ntflx, v-skype, v-vimeo, v-ytube
},
x tick label style={
font=\fontsize{3.5}{1}\selectfont, rotate=45, anchor=north east
},
y tick label style={font=\tiny},
xtick style={draw=none},
ytick style={draw=none},
]

\addplot+[
    blue!90!black,fill=blue!35!white,
    error bars/.cd,
    y dir=both,
    y explicit relative,
    error mark options={
      rotate=90,
      mark size=0.9pt,
      line width=0.5pt
    }
] coordinates {
    (a-fb,13.991)            +- (0, 99.651)
    (a-hng,13.688)           +- (0, 106.888)
    (a-skype,13.009)         +- (0, 68.260)
    (a-sptfy,64.942)         +- (0, 9.523)
    (a-vpbust,143.831)       +- (0, 32.816)
    (c-aim,6.761)            +- (0, 5.437)
    (c-fb,11.756)            +- (0, 6.534)
    (c-gmail,14.201)        +- (0, 6.006)
    (c-hng,16.410)           +- (0, 7.044)
    (c-icq,6.201)            +- (0, 5.793)
    (c-skype,5.905)          +- (0, 15.093)
    (e-email,4.719)          +- (0, 5.812)
    (ft-ftps,3514.885)       +- (0, 9.299)
    (ft-scp,31.816)          +- (0, 27.518)
    (ft-sftp,1321.473)       +- (0, 5.655)
    (ft-skype,13.751)        +- (0, 110.005)
    (p2p-torr,36.020)        +- (0, 3.111)
    (v-fb,866.797)           +- (0, 8.987)
    (v-hng,635.429)          +- (0, 25.167)
    (v-ntflx,407.468)        +- (0, 10.198)
    (v-skype,1063.136)       +- (0, 10.629)
    (v-vimeo,144.650)        +- (0, 10.850)
    (v-ytube,122.489)        +- (0, 15.438)
};
\addplot+[
    blue!90!black,fill=blue!60!white,
    error bars/.cd,
    y dir=both,
    y explicit relative,
    error mark options={
      rotate=90,
      mark size=0.9pt,
      line width=0.5pt
    }
] coordinates {
    (a-fb,9.255)             +- (0, 149.481)
    (a-hng,9.701)            +- (0, 149.958)
    (a-skype,9.725)          +- (0, 90.875)
    (a-sptfy,75.577)         +- (0, 9.920)
    (a-vpbust,143.018)       +- (0, 33.365)
    (c-aim,4.725)            +- (0, 8.032)
    (c-fb,9.860)             +- (0, 7.661)
    (c-gmail,13.116)        +- (0, 5.412)
    (c-hng,14.002)           +- (0, 7.715)
    (c-icq,3.296)            +- (0, 9.007)
    (c-skype,2.006)          +- (0, 45.021)
    (e-email,1.367)          +- (0, 11.790)
    (ft-ftps,5686.356)       +- (0, 12.133)
    (ft-scp,41.067)          +- (0, 33.902)
    (ft-sftp,2515.554)       +- (0, 5.374)
    (ft-skype,10.079)        +- (0, 148.942)
    (p2p-torr,66.437)        +- (0, 3.253)
    (v-fb,857.828)           +- (0, 9.043)
    (v-hng,669.343)          +- (0, 22.813)
    (v-ntflx,392.684)        +- (0, 9.812)
    (v-skype,1523.505)       +- (0, 9.913)
    (v-vimeo,181.345)        +- (0, 11.056)
    (v-ytube,150.609)        +- (0, 15.970)
};

\addplot+[
    green!20!black,fill=green!35!white,
    error bars/.cd,
    y dir=both,
    y explicit relative,
    error mark options={
      rotate=90,
      mark size=0.9pt,
      line width=0.5pt
    }
] coordinates {
    (a-fb,5.571)             +- (0, 13.640)
    (a-hng,56.852)           +- (0, 50.276)
    (a-skype,407.197)        +- (0, 21.131)
    (a-sptfy,341.417)        +- (0, 10.312)
    (a-vpbust,224.981)       +- (0, 28.007)
    (c-aim,37.485)           +- (0, 3.416)
    (c-fb,6.668)             +- (0, 17.402)
    (c-gmail,0.000)          +- (0, 0.000)
    (c-hng,6.987)            +- (0, 15.262)
    (c-icq,125.419)          +- (0, 3.732)
    (c-skype,127.268)        +- (0, 2.787)
    (e-email,33.040)         +- (0, 6.821)
    (ft-ftps,767.551)        +- (0, 4.165)
    (ft-scp,0.000)           +- (0, 0.000)
    (ft-sftp,2211.300)       +- (0, 4.385)
    (ft-skype,47.497)        +- (0, 4.829)
    (p2p-torr,319.274)       +- (0, 1.754)
    (v-fb,0.000)             +- (0, 0.000)
    (v-hng,0.000)            +- (0, 0.000)
    (v-ntflx,1663.693)       +- (0, 13.139)
    (v-skype,0.000)          +- (0, 0.000)
    (v-vimeo,332.181)        +- (0, 7.388)
    (v-ytube,367.907)        +- (0, 3.586)
};
\addplot+[
    green!20!black,fill=green!25!gray,
    error bars/.cd,
    y dir=both,
    y explicit relative,
    error mark options={
      rotate=90,
      mark size=0.9pt,
      line width=0.5pt
    }
] coordinates {
    (a-fb,5.701)             +- (0, 13.236)
    (a-hng,57.083)           +- (0, 49.018)
    (a-skype,406.230)        +- (0, 21.164)
    (a-sptfy,624.375)        +- (0, 10.594)
    (a-vpbust,222.498)       +- (0, 28.291)
    (c-aim,36.909)           +- (0, 3.430)
    (c-fb,7.418)             +- (0, 18.030)
    (c-gmail,0.000)         +- (0, 0.000)
    (c-hng,9.335)            +- (0, 16.293)
    (c-icq,128.677)          +- (0, 3.763)
    (c-skype,132.964)        +- (0, 2.843)
    (e-email,38.897)         +- (0, 7.229)
    (ft-ftps,738.386)        +- (0, 4.326)
    (ft-scp,0.000)           +- (0, 0.000)
    (ft-sftp,1491.433)       +- (0, 3.675)
    (ft-skype,38.941)        +- (0, 5.606)
    (p2p-torr,551.027)       +- (0, 1.862)
    (v-fb,0.000)             +- (0, 0.000)
    (v-hng,0.000)            +- (0, 0.000)
    (v-ntflx,3196.743)       +- (0, 13.239)
    (v-skype,0.000)          +- (0, 0.000)
    (v-vimeo,1986.486)       +- (0, 7.179)
    (v-ytube,605.715)        +- (0, 3.806)
};

\end{axis}

\end{tikzpicture}

%% file: assets/tikz/iscx/iscx_avg_bytes.tikz

\begin{tikzpicture}
 
\begin{axis} [ybar=0.1pt,
height = 4cm, width = 12cm,
bar width = 2pt,
ymin = 1,
ymode=log,
title=\textcolor{blue}{(\ref{app:fig:iscx_avg_bytes})} AVG Bytes x Session,every axis title/.style={below right,at={(0,1)},draw=gray,fill=black!5,font=\fontsize{6}{6}\selectfont},
log basis y={10},
ytick distance=10,
ymajorgrids=true,
enlarge x limits=0.035,
xticklabel shift={-5pt},
yticklabel shift={-2pt},
symbolic x coords={
a-fb,a-hng,a-skype,a-sptfy,a-vpbust,c-aim,c-fb, c-gmail, c-hng, c-icq, c-skype, e-email, ft-ftps, ft-scp, ft-sftp, ft-skype, p2p-torr, v-fb, v-hng, v-ntflx, v-skype, v-vimeo, v-ytube
},
xtick={
a-fb,a-hng,a-skype,a-sptfy,a-vpbust,c-aim,c-fb, c-gmail, c-hng, c-icq, c-skype, e-email, ft-ftps, ft-scp, ft-sftp, ft-skype, p2p-torr, v-fb, v-hng, v-ntflx, v-skype, v-vimeo, v-ytube
},
x tick label style={
font=\fontsize{3.5}{1}\selectfont, rotate=45, anchor=north east
},
y tick label style={font=\tiny},
xtick style={draw=none},
ytick style={draw=none},
]

\addplot+[
    blue!90!black,fill=blue!35!white,
    error bars/.cd,
    y dir=both,
    y explicit relative,
    error mark options={
      rotate=90,
      mark size=0.9pt,
      line width=0.5pt
    }
] coordinates {
    (a-fb,1930.694)          +- (0, 109.723)
    (a-hng,1991.158)         +- (0, 111.332)
    (a-skype,1595.988)       +- (0, 74.233)
    (a-sptfy,5856.962)       +- (0, 7.693)
    (a-vpbust,17911.570)     +- (0, 32.912)
    (c-aim,844.614)          +- (0, 5.673)
    (c-fb,3246.100)          +- (0, 8.200)
    (c-gmail,8140.141)       +- (0, 8.217)
    (c-hng,5963.894)         +- (0, 9.873)
    (c-icq,629.145)          +- (0, 6.796)
    (c-skype,1046.309)       +- (0, 27.156)
    (e-email,484.656)        +- (0, 5.782)
    (ft-ftps,399263.000)     +- (0, 7.945)
    (ft-scp,425968.400)      +- (0, 34.639)
    (ft-sftp,3943948.000)    +- (0, 7.369)
    (ft-skype,9867.128)      +- (0, 161.632)
    (p2p-torr,2352.249)      +- (0, 2.841)
    (v-fb,795630.200)        +- (0, 10.657)
    (v-hng,550581.900)       +- (0, 27.885)
    (v-ntflx,972268.100)     +- (0, 14.128)
    (v-skype,691089.100)     +- (0, 13.421)
    (v-vimeo,254803.000)     +- (0, 20.536)
    (v-ytube,9098.079)       +- (0, 13.986)
};
\addplot+[
    blue!90!black,fill=blue!60!white,
    error bars/.cd,
    y dir=both,
    y explicit relative,
    error mark options={
      rotate=90,
      mark size=0.9pt,
      line width=0.5pt
    }
] coordinates {
    (a-fb,1051.552)          +- (0, 143.963)
    (a-hng,1543.925)         +- (0, 141.968)
    (a-skype,1445.864)       +- (0, 85.874)
    (a-sptfy,242428.800)     +- (0, 10.475)
    (a-vpbust,18072.830)     +- (0, 32.975)
    (c-aim,2139.717)         +- (0, 14.103)
    (c-fb,3846.040)          +- (0, 7.467)
    (c-gmail,3030.989)       +- (0, 5.818)
    (c-hng,7006.203)         +- (0, 9.799)
    (c-icq,526.434)          +- (0, 8.739)
    (c-skype,576.120)        +- (0, 42.144)
    (e-email,194.859)        +- (0, 13.127)
    (ft-ftps,6776720.000)    +- (0, 13.458)
    (ft-scp,72286.130)       +- (0, 71.775)
    (ft-sftp,1904774.000)    +- (0, 7.067)
    (ft-skype,5186.843)      +- (0, 126.786)
    (p2p-torr,90211.700)     +- (0, 3.349)
    (v-fb,798199.100)        +- (0, 10.574)
    (v-hng,583161.000)       +- (0, 24.134)
    (v-ntflx,733959.400)     +- (0, 12.983)
    (v-skype,1611515.000)    +- (0, 11.684)
    (v-vimeo,790839.200)     +- (0, 13.022)
    (v-ytube,481206.000)     +- (0, 16.418)
};

\addplot+[
    green!20!black,fill=green!35!white,
    error bars/.cd,
    y dir=both,
    y explicit relative,
    error mark options={
      rotate=90,
      mark size=0.9pt,
      line width=0.5pt
    }
] coordinates {
    (a-fb,1776.701)          +- (0, 25.929)
    (a-hng,8684.775)         +- (0, 45.666)
    (a-skype,52595.070)      +- (0, 21.166)
    (a-sptfy,22350.080)      +- (0, 8.782)
    (a-vpbust,28081.130)     +- (0, 28.048)
    (c-aim,5171.333)         +- (0, 3.775)
    (c-fb,2201.439)          +- (0, 22.275)
    (c-gmail,0.000)          +- (0, 0.000)
    (c-hng,1972.369)         +- (0, 22.528)
    (c-icq,12438.710)        +- (0, 3.937)
    (c-skype,16488.380)      +- (0, 2.726)
    (e-email,6946.513)       +- (0, 11.906)
    (ft-ftps,725859.900)     +- (0, 5.443)
    (ft-scp,0.000)           +- (0, 0.000)
    (ft-sftp,2638291.000)    +- (0, 4.879)
    (ft-skype,9977.748)      +- (0, 5.723)
    (p2p-torr,18196.690)     +- (0, 1.734)
    (v-fb,0.000)             +- (0, 0.000)
    (v-hng,0.000)            +- (0, 0.000)
    (v-ntflx,94745.870)      +- (0, 12.423)
    (v-skype,0.000)          +- (0, 0.000)
    (v-vimeo,19985.830)      +- (0, 6.580)
    (v-ytube,23582.370)      +- (0, 3.311)
};
\addplot+[
    green!20!black,fill=green!25!gray,
    error bars/.cd,
    y dir=both,
    y explicit relative,
    error mark options={
      rotate=90,
      mark size=0.9pt,
      line width=0.5pt
    }
] coordinates {
    (a-fb,2898.721)          +- (0, 13.717)
    (a-hng,9348.545)         +- (0, 40.964)
    (a-skype,41860.480)      +- (0, 20.817)
    (a-sptfy,823356.800)     +- (0, 10.752)
    (a-vpbust,28610.210)     +- (0, 27.508)
    (c-aim,4816.970)         +- (0, 3.615)
    (c-fb,2728.565)          +- (0, 17.967)
    (c-gmail,0.000)          +- (0, 0.000)
    (c-hng,4993.362)         +- (0, 20.419)
    (c-icq,12714.520)        +- (0, 3.853)
    (c-skype,12445.040)      +- (0, 2.694)
    (e-email,19186.320)      +- (0, 14.631)
    (ft-ftps,643869.900)     +- (0, 6.076)
    (ft-scp,0.000)           +- (0, 0.000)
    (ft-sftp,644419.800)     +- (0, 4.346)
    (ft-skype,6589.300)      +- (0, 7.310)
    (p2p-torr,743522.400)    +- (0, 1.871)
    (v-fb,0.000)             +- (0, 0.000)
    (v-hng,0.000)            +- (0, 0.000)
    (v-ntflx,4332846.000)    +- (0, 13.271)
    (v-skype,0.000)          +- (0, 0.000)
    (v-vimeo,2690673.000)    +- (0, 7.204)
    (v-ytube,812392.600)     +- (0, 3.847)
};

\end{axis}

\end{tikzpicture}

%% file: assets/tikz/iscx/iscx_avg_duration_s.tikz

\begin{tikzpicture}
 
\begin{axis} [ybar=0.1pt,
height = 4cm, width = 12cm,
bar width = 2pt,
ymin = 1,
ymode=log,
title=\textcolor{blue}{(\ref{app:fig:iscx_avg_dur_s})} AVG duration (sec) x Session,every axis title/.style={below right,at={(0,1)},draw=gray,fill=black!5,font=\fontsize{6}{6}\selectfont},
log basis y={10},
ytick distance=10,
ymajorgrids=true,
enlarge x limits=0.035,
xticklabel shift={-5pt},
yticklabel shift={-2pt},
symbolic x coords={
a-fb,a-hng,a-skype,a-sptfy,a-vpbust,c-aim,c-fb, c-gmail, c-hng, c-icq, c-skype, e-email, ft-ftps, ft-scp, ft-sftp, ft-skype, p2p-torr, v-fb, v-hng, v-ntflx, v-skype, v-vimeo, v-ytube
},
xtick={
a-fb,a-hng,a-skype,a-sptfy,a-vpbust,c-aim,c-fb, c-gmail, c-hng, c-icq, c-skype, e-email, ft-ftps, ft-scp, ft-sftp, ft-skype, p2p-torr, v-fb, v-hng, v-ntflx, v-skype, v-vimeo, v-ytube
},
x tick label style={
font=\fontsize{3.5}{1}\selectfont, rotate=45, anchor=north east
},
y tick label style={font=\tiny},
xtick style={draw=none},
ytick style={draw=none},
]

\addplot+[
    blue!90!black,fill=blue!35!white,
    error bars/.cd,
    y dir=both,
    y explicit relative,
    error mark options={
      rotate=90,
      mark size=0.9pt,
      line width=0.5pt
    }
] coordinates {
    (a-fb,508.912)           +- (0, 2.308)
    (a-hng,356.473)          +- (0, 2.629)
    (a-skype,105.657)        +- (0, 3.492)
    (a-sptfy,176.758)        +- (0, 2.034)
    (a-vpbust,48.464)        +- (0, 6.014)
    (c-aim,31.725)           +- (0, 3.548)
    (c-fb,25.219)            +- (0, 3.727)
    (c-gmail,42.448)         +- (0, 3.234)
    (c-hng,26.246)           +- (0, 4.445)
    (c-icq,32.955)           +- (0, 3.725)
    (c-skype,46.205)         +- (0, 3.939)
    (e-email,52.339)         +- (0, 6.133)
    (ft-ftps,67.334)         +- (0, 2.719)
    (ft-scp,15.215)          +- (0, 4.869)
    (ft-sftp,18.125)         +- (0, 2.577)
    (ft-skype,272.952)       +- (0, 2.653)
    (p2p-torr,10.902)        +- (0, 4.221)
    (v-fb,87.339)            +- (0, 1.726)
    (v-hng,32.073)           +- (0, 5.481)
    (v-ntflx,104.675)        +- (0, 1.038)
    (v-skype,60.196)         +- (0, 3.116)
    (v-vimeo,52.487)         +- (0, 2.048)
    (v-ytube,47.202)         +- (0, 2.334)
};
\addplot+[
    blue!90!black,fill=blue!60!white,
    error bars/.cd,
    y dir=both,
    y explicit relative,
    error mark options={
      rotate=90,
      mark size=0.9pt,
      line width=0.5pt
    }
] coordinates {
    (a-fb,3.539)             +- (0, 31.620)
    (a-hng,3.524)            +- (0, 26.826)
    (a-skype,2.017)          +- (0, 24.898)
    (a-sptfy,119.690)        +- (0, 2.355)
    (a-vpbust,14.264)        +- (0, 11.460)
    (c-aim,25.067)           +- (0, 4.041)
    (c-fb,19.503)            +- (0, 4.129)
    (c-gmail,40.629)         +- (0, 3.297)
    (c-hng,21.747)           +- (0, 4.958)
    (c-icq,27.981)           +- (0, 4.073)
    (c-skype,2.052)          +- (0, 17.577)
    (e-email,20.488)         +- (0, 13.324)
    (ft-ftps,52.366)         +- (0, 3.074)
    (ft-scp,2.820)           +- (0, 10.311)
    (ft-sftp,14.344)         +- (0, 3.065)
    (ft-skype,2.625)         +- (0, 27.776)
    (p2p-torr,8.182)         +- (0, 4.745)
    (v-fb,70.622)            +- (0, 1.986)
    (v-hng,23.970)           +- (0, 6.172)
    (v-ntflx,96.500)         +- (0, 1.093)
    (v-skype,48.536)         +- (0, 3.498)
    (v-vimeo,40.410)         +- (0, 2.287)
    (v-ytube,35.100)         +- (0, 2.374)
};

\addplot+[
    green!20!black,fill=green!35!white,
    error bars/.cd,
    y dir=both,
    y explicit relative,
    error mark options={
      rotate=90,
      mark size=0.9pt,
      line width=0.5pt
    }
] coordinates {
    (a-fb,107.348)           +- (0, 5.326)
    (a-hng,27.724)           +- (0, 9.082)
    (a-skype,458.794)        +- (0, 2.305)
    (a-sptfy,296.524)        +- (0, 2.984)
    (a-vpbust,34.590)        +- (0, 9.216)
    (c-aim,307.443)          +- (0, 1.326)
    (c-fb,9.185)             +- (0, 8.599)
    (c-gmail,0.000)          +- (0, 0.000)
    (c-hng,8.660)            +- (0, 7.864)
    (c-icq,320.967)          +- (0, 1.615)
    (c-skype,217.022)        +- (0, 1.636)
    (e-email,23.091)         +- (0, 5.776)
    (ft-ftps,32.407)         +- (0, 2.149)
    (ft-scp,0.000)           +- (0, 0.000)
    (ft-sftp,77.795)         +- (0, 1.286)
    (ft-skype,170.395)       +- (0, 3.249)
    (p2p-torr,8.686)         +- (0, 2.218)
    (v-fb,0.000)             +- (0, 0.000)
    (v-hng,0.000)            +- (0, 0.000)
    (v-ntflx,142.694)        +- (0, 1.988)
    (v-skype,0.000)          +- (0, 0.000)
    (v-vimeo,166.764)        +- (0, 1.954)
    (v-ytube,122.145)        +- (0, 2.994)
};
\addplot+[
    green!20!black,fill=green!25!gray,
    error bars/.cd,
    y dir=both,
    y explicit relative,
    error mark options={
      rotate=90,
      mark size=0.9pt,
      line width=0.5pt
    }
] coordinates {
(a-fb,98.532)            +- (0, 5.508)
(a-hng,24.909)           +- (0, 9.205)
(a-skype,419.654)        +- (0, 2.428)
(a-sptfy,202.200)        +- (0, 3.397)
(a-vpbust,13.159)        +- (0, 15.876)
(c-aim,306.786)          +- (0, 1.327)
(c-fb,7.312)             +- (0, 9.186)
(c-gmail,0.000)          +- (0, 0.000)
(c-hng,7.992)            +- (0, 7.949)
(c-icq,320.906)          +- (0, 1.615)
(c-skype,216.954)        +- (0, 1.636)
(e-email,16.627)         +- (0, 6.494)
(ft-ftps,27.846)         +- (0, 2.181)
(ft-scp,0.000)           +- (0, 0.000)
(ft-sftp,75.087)         +- (0, 1.346)
(ft-skype,139.207)       +- (0, 3.412)
(p2p-torr,6.868)         +- (0, 2.332)
(v-fb,0.000)             +- (0, 0.000)
(v-hng,0.000)            +- (0, 0.000)
(v-ntflx,125.404)        +- (0, 1.921)
(v-skype,0.000)          +- (0, 0.000)
(v-vimeo,133.750)        +- (0, 2.090)
(v-ytube,97.588)         +- (0, 2.805)

};

\end{axis}

\end{tikzpicture}

%% file: assets/tikz/iscx/iscx_avg_iat_ms.tikz

\begin{tikzpicture}
 
\begin{axis} [ybar=0.1pt,
height = 4cm, width = 12cm,
bar width = 2pt,
ymin = 1,
ymode=log,
title=\textcolor{blue}{(\ref{app:fig:iscx_avg_iat_ms})} AVG IAT (ms) x Session,every axis title/.style={below right,at={(0,1)},draw=gray,fill=black!5,font=\fontsize{6}{6}\selectfont},
log basis y={10},
ytick distance=10,
ymajorgrids=true,
enlarge x limits=0.035,
xticklabel shift={-5pt},
yticklabel shift={-2pt},
symbolic x coords={
a-fb,a-hng,a-skype,a-sptfy,a-vpbust,c-aim,c-fb, c-gmail, c-hng, c-icq, c-skype, e-email, ft-ftps, ft-scp, ft-sftp, ft-skype, p2p-torr, v-fb, v-hng, v-ntflx, v-skype, v-vimeo, v-ytube
},
xtick={
a-fb,a-hng,a-skype,a-sptfy,a-vpbust,c-aim,c-fb, c-gmail, c-hng, c-icq, c-skype, e-email, ft-ftps, ft-scp, ft-sftp, ft-skype, p2p-torr, v-fb, v-hng, v-ntflx, v-skype, v-vimeo, v-ytube
},
x tick label style={
font=\fontsize{3.5}{1}\selectfont, rotate=45, anchor=north east
},
y tick label style={font=\tiny},
xtick style={draw=none},
ytick style={draw=none},
]

\addplot+[
    blue!90!black,fill=blue!35!white,
    error bars/.cd,
    y dir=both,
    y explicit relative,
    error mark options={
      rotate=90,
      mark size=0.9pt,
      line width=0.5pt
    }
] coordinates {
    (a-fb,47349.771)         +- (0, 4.430)
    (a-hng,33908.469)        +- (0, 4.889)
    (a-skype,10490.121)      +- (0, 6.025)
    (a-sptfy,3351.716)       +- (0, 7.268)
    (a-vpbust,432.145)       +- (0, 30.983)
    (c-aim,5316.130)         +- (0, 2.716)
    (c-fb,2431.098)          +- (0, 3.969)
    (c-gmail,3149.456)       +- (0, 2.344)
    (c-hng,1809.951)         +- (0, 6.722)
    (c-icq,5992.328)         +- (0, 2.939)
    (c-skype,9083.368)       +- (0, 3.602)
    (e-email,14264.563)      +- (0, 7.962)
    (ft-ftps,22.179)         +- (0, 59.135)
    (ft-scp,595.603)         +- (0, 11.913)
    (ft-sftp,16.775)         +- (0, 46.758)
    (ft-skype,25837.470)     +- (0, 4.907)
    (p2p-torr,430.003)       +- (0, 14.711)
    (v-fb,113.695)           +- (0, 20.192)
    (v-hng,58.140)           +- (0, 54.821)
    (v-ntflx,308.902)        +- (0, 12.830)
    (v-skype,64.853)         +- (0, 30.849)
    (v-vimeo,430.672)        +- (0, 11.181)
    (v-ytube,466.198)        +- (0, 14.164)
};
\addplot+[
    blue!90!black,fill=blue!60!white,
    error bars/.cd,
    y dir=both,
    y explicit relative,
    error mark options={
      rotate=90,
      mark size=0.9pt,
      line width=0.5pt
    }
] coordinates {
    (a-fb,439.597)           +- (0, 31.724)
    (a-hng,554.766)          +- (0, 55.939)
    (a-skype,234.585)        +- (0, 21.959)
    (a-sptfy,1896.556)       +- (0, 8.367)
    (a-vpbust,119.950)       +- (0, 27.729)
    (c-aim,5955.905)         +- (0, 2.537)
    (c-fb,2190.439)          +- (0, 3.463)
    (c-gmail,3249.256)       +- (0, 2.227)
    (c-hng,1726.213)         +- (0, 7.153)
    (c-icq,9595.923)         +- (0, 2.425)
    (c-skype,1112.052)       +- (0, 5.532)
    (e-email,21889.717)      +- (0, 8.900)
    (ft-ftps,10.760)         +- (0, 84.948)
    (ft-scp,78.143)          +- (0, 10.508)
    (ft-sftp,6.498)          +- (0, 61.752)
    (ft-skype,320.674)       +- (0, 42.037)
    (p2p-torr,176.839)       +- (0, 22.360)
    (v-fb,96.364)            +- (0, 25.573)
    (v-hng,40.642)           +- (0, 60.373)
    (v-ntflx,287.277)        +- (0, 10.712)
    (v-skype,36.581)         +- (0, 41.389)
    (v-vimeo,251.768)        +- (0, 8.835)
    (v-ytube,263.509)        +- (0, 9.049)
};

\addplot+[
    green!20!black,fill=green!35!white,
    error bars/.cd,
    y dir=both,
    y explicit relative,
    error mark options={
      rotate=90,
      mark size=0.9pt,
      line width=0.5pt
    }
] coordinates {
    (a-fb,22129.574)         +- (0, 4.603)
    (a-hng,740.168)          +- (0, 43.284)
    (a-skype,1622.979)       +- (0, 30.360)
    (a-sptfy,890.273)        +- (0, 6.724)
    (a-vpbust,174.048)       +- (0, 43.336)
    (c-aim,10311.283)        +- (0, 4.410)
    (c-fb,1588.760)          +- (0, 8.041)
    (c-gmail,0.000)          +- (0, 0.000)
    (c-hng,1793.504)         +- (0, 13.344)
    (c-icq,2955.041)         +- (0, 7.472)
    (c-skype,2060.748)       +- (0, 9.066)
    (e-email,727.654)        +- (0, 6.224)
    (ft-ftps,46.760)         +- (0, 23.544)
    (ft-scp,0.000)           +- (0, 0.000)
    (ft-sftp,39.432)         +- (0, 23.810)
    (ft-skype,4222.039)      +- (0, 10.401)
    (p2p-torr,32.831)        +- (0, 22.299)
    (v-fb,0.000)             +- (0, 0.000)
    (v-hng,0.000)            +- (0, 0.000)
    (v-ntflx,93.554)         +- (0, 16.748)
    (v-skype,0.000)          +- (0, 0.000)
    (v-vimeo,545.505)        +- (0, 8.581)
    (v-ytube,346.430)        +- (0, 9.163)
};
\addplot+[
    green!20!black,fill=green!25!gray,
    error bars/.cd,
    y dir=both,
    y explicit relative,
    error mark options={
      rotate=90,
      mark size=0.9pt,
      line width=0.5pt
    }
] coordinates {
    (a-fb,20800.527)         +- (0, 5.483)
    (a-hng,686.423)          +- (0, 46.464)
    (a-skype,1519.841)       +- (0, 32.207)
    (a-sptfy,332.870)        +- (0, 6.791)
    (a-vpbust,60.663)        +- (0, 36.580)
    (c-aim,10888.227)        +- (0, 4.692)
    (c-fb,1197.870)          +- (0, 11.264)
    (c-gmail,0.000)          +- (0, 0.000)
    (c-hng,1283.364)         +- (0, 16.451)
    (c-icq,2967.505)         +- (0, 8.477)
    (c-skype,1996.602)       +- (0, 9.485)
    (e-email,436.821)        +- (0, 2.128)
    (ft-ftps,40.471)         +- (0, 17.548)
    (ft-scp,0.000)           +- (0, 0.000)
    (ft-sftp,54.637)         +- (0, 18.054)
    (ft-skype,4393.960)      +- (0, 11.259)
    (p2p-torr,14.287)        +- (0, 30.339)
    (v-fb,0.000)             +- (0, 0.000)
    (v-hng,0.000)            +- (0, 0.000)
    (v-ntflx,44.044)         +- (0, 25.194)
    (v-skype,0.000)          +- (0, 0.000)
    (v-vimeo,73.139)         +- (0, 18.266)
    (v-ytube,170.281)        +- (0, 10.262)
};

\end{axis}

\end{tikzpicture}

%% file: assets/tikz/iscx/iscx_protocol_dist_nonvpn.tikz
\begin{tikzpicture}
 
\begin{axis} [ybar=0.1pt,
height = 3cm, width = 12cm,
bar width = 2pt,
title=\textcolor{blue}{(\ref{app:fig:iscx_proto_nonvpn})} Protocol x Session (Non-VPN),every axis title/.style={below left,at={(1,1)},draw=gray,fill=black!5,font=\fontsize{5}{6}\selectfont},
label style={font=\small},
ymin = 1,
ymode=log,
log basis y={10},
ytick distance=10,
ymajorgrids=true,
enlarge x limits=0.02,
xticklabel shift={-5pt},
yticklabel shift={-2pt},
symbolic x coords={
a-fb,a-hng,a-skype,a-sptfy,a-vpbust,c-aim,c-fb, c-gmail, c-hng, c-icq, c-skype, e-email, ft-ftps, ft-scp, ft-sftp, ft-skype, p2p-torr, v-fb, v-hng, v-ntflx, v-skype, v-vimeo, v-ytube
},
xtick={
a-fb,a-hng,a-skype,a-sptfy,a-vpbust,c-aim,c-fb, c-gmail, c-hng, c-icq, c-skype, e-email, ft-ftps, ft-scp, ft-sftp, ft-skype, p2p-torr, v-fb, v-hng, v-ntflx, v-skype, v-vimeo, v-ytube
},
x tick label style={
font=\fontsize{3.5}{1}\selectfont, rotate=45, anchor=north east
},
y tick label style={font=\tiny},
xtick style={draw=none},
ytick style={draw=none},
]

\addplot+[
    gray!90!black,fill=gray!60!white,
] coordinates {
    (a-fb,87.000)
    (a-hng,74.000)
    (a-skype,47.000)
    (a-sptfy,75.000)
    (a-vpbust,19.000)
    (c-aim,0.000)
    (c-fb,0.000)
    (c-gmail,1.000)
    (c-hng,0.000)
    (c-icq,0.000)
    (c-skype,1.000)
    (e-email,2.000)
    (ft-ftps,18.000)
    (ft-scp,38.000)
    (ft-sftp,7.000)
    (ft-skype,60.000)
    (p2p-torr,22.000)
    (v-fb,15.000)
    (v-hng,19.000)
    (v-ntflx,42.000)
    (v-skype,28.000)
    (v-vimeo,19.000)
    (v-ytube,41.000)
};
\addplot+[
    red!50!gray,fill=red!35!white,
] coordinates {
    (a-fb,198.000)
    (a-hng,193.000)
    (a-skype,112.000)
    (a-sptfy,0.000)
    (a-vpbust,8.000)
    (c-aim,0.000)
    (c-fb,0.000)
    (c-gmail,0.000)
    (c-hng,0.000)
    (c-icq,1.000)
    (c-skype,74.000)
    (e-email,50.000)
    (ft-ftps,4.000)
    (ft-scp,35.000)
    (ft-sftp,1.000)
    (ft-skype,173.000)
    (p2p-torr,2.000)
    (v-fb,2.000)
    (v-hng,1.000)
    (v-ntflx,0.000)
    (v-skype,0.000)
    (v-vimeo,0.000)
    (v-ytube,5.000)
};

\addplot+[
    green!90!black,fill=green!35!white,
] coordinates {
    (a-fb,273.000)
    (a-hng,275.000)
    (a-skype,358.000)
    (a-sptfy,123.000)
    (a-vpbust,89.000)
    (c-aim,45.000)
    (c-fb,78.000)
    (c-gmail,72.000)
    (c-hng,46.000)
    (c-icq,41.000)
    (c-skype,160.000)
    (e-email,239.000)
    (ft-ftps,180.000)
    (ft-scp,1175.000)
    (ft-sftp,101.000)
    (ft-skype,334.000)
    (p2p-torr,301.000)
    (v-fb,121.000)
    (v-hng,124.000)
    (v-ntflx,270.000)
    (v-skype,367.000)
    (v-vimeo,282.000)
    (v-ytube,467.000)
};

\addplot+[
    blue!90!black,fill=blue!35!white,
] coordinates {
    (a-fb,79197.000)
    (a-hng,78693.000)
    (a-skype,38159.000)
    (a-sptfy,95.000)
    (a-vpbust,2821.000)
    (c-aim,369.000)
    (c-fb,423.000)
    (c-gmail,375.000)
    (c-hng,388.000)
    (c-icq,391.000)
    (c-skype,8393.000)
    (e-email,7065.000)
    (ft-ftps,639.000)
    (ft-scp,10128.000)
    (ft-sftp,115.000)
    (ft-skype,56587.000)
    (p2p-torr,734.000)
    (v-fb,291.000)
    (v-hng,1381.000)
    (v-ntflx,62.000)
    (v-skype,193.000)
    (v-vimeo,148.000)
    (v-ytube,410.000)
};

\addplot+[
    yellow!60!black,fill=yellow!35!white,
] coordinates {
    (a-fb,183.000)
    (a-hng,181.000)
    (a-skype,128.000)
    (a-sptfy,0.000)
    (a-vpbust,0.000)
    (c-aim,0.000)
    (c-fb,0.000)
    (c-gmail,0.000)
    (c-hng,0.000)
    (c-icq,0.000)
    (c-skype,194.000)
    (e-email,154.000)
    (ft-ftps,0.000)
    (ft-scp,88.000)
    (ft-sftp,0.000)
    (ft-skype,153.000)
    (p2p-torr,0.000)
    (v-fb,0.000)
    (v-hng,0.000)
    (v-ntflx,0.000)
    (v-skype,0.000)
    (v-vimeo,0.000)
    (v-ytube,0.000)
};

\end{axis}

\end{tikzpicture}

%% file: assets/tikz/iscx/iscx_protocol_dist.tikz
\begin{tikzpicture}
 
\begin{axis} [ybar=0.1pt,
height = 3cm, width = 12cm,
bar width = 2pt,
title=\textcolor{blue}{(\ref{app:fig:iscx_proto})} Protocol x Session (VPN),every axis title/.style={below left,at={(1,1)},draw=gray,fill=black!5,font=\fontsize{5}{6}\selectfont},
label style={font=\small},
ymin = 1,
ymode=log,
log basis y={10},
ytick distance=10,
ymajorgrids=true,
enlarge x limits=0.02,
xticklabel shift={-5pt},
yticklabel shift={-2pt},
symbolic x coords={
a-fb,a-hng,a-skype,a-sptfy,a-vpbust,c-aim,c-fb, c-gmail, c-hng, c-icq, c-skype, e-email, ft-ftps, ft-scp, ft-sftp, ft-skype, p2p-torr, v-fb, v-hng, v-ntflx, v-skype, v-vimeo, v-ytube
},
xtick={
a-fb,a-hng,a-skype,a-sptfy,a-vpbust,c-aim,c-fb, c-gmail, c-hng, c-icq, c-skype, e-email, ft-ftps, ft-scp, ft-sftp, ft-skype, p2p-torr, v-fb, v-hng, v-ntflx, v-skype, v-vimeo, v-ytube
},
x tick label style={
font=\fontsize{3.5}{1}\selectfont, rotate=45, anchor=north east
},
y tick label style={font=\tiny},
xtick style={draw=none},
ytick style={draw=none},
]

\addplot+[
    gray!90!black,fill=gray!60!white,
] coordinates {
    (a-fb,4.000)            
    (a-hng,7.000)           
    (a-skype,11.000)        
    (a-sptfy,1.000)         
    (a-vpbust,1.000)        
    (c-aim,1.000)           
    (c-fb,3.000)            
    (c-gmail,0.000)         
    (c-hng,3.000)           
    (c-icq,0.000)           
    (c-skype,2.000)         
    (e-email,2.000)         
    (ft-ftps,2.000)         
    (ft-scp,0.000)          
    (ft-sftp,2.000)         
    (ft-skype,21.000)       
    (p2p-torr,10.000)       
    (v-fb,0.000)            
    (v-hng,0.000)           
    (v-ntflx,6.000)         
    (v-skype,0.000)         
    (v-vimeo,2.000)         
    (v-ytube,1.000)         
};
\addplot+[
    red!50!gray,fill=red!35!white,
] coordinates {
    (a-fb,0.000)    
    (a-hng,0.000)   
    (a-skype,0.000) 
    (a-sptfy,0.000) 
    (a-vpbust,1.000)
    (c-aim,0.000)   
    (c-fb,0.000)    
    (c-gmail,0.000) 
    (c-hng,0.000)   
    (c-icq,0.000)   
    (c-skype,0.000) 
    (e-email,0.000) 
    (ft-ftps,0.000) 
    (ft-scp,0.000)  
    (ft-sftp,0.000) 
    (ft-skype,0.000)
    (p2p-torr,0.000)
    (v-fb,0.000)    
    (v-hng,0.000)   
    (v-ntflx,0.000) 
    (v-skype,0.000) 
    (v-vimeo,0.000) 
    (v-ytube,0.000) 
};

\addplot+[
    green!90!black,fill=green!35!white,
] coordinates {
    (a-fb,139.000)  
    (a-hng,264.000) 
    (a-skype,156.000)  
    (a-sptfy,51.000)
    (a-vpbust,36.000)  
    (c-aim,13.000)  
    (c-fb,59.000)   
    (c-gmail,0.000) 
    (c-hng,97.000)  
    (c-icq,8.000)   
    (c-skype,19.000)
    (e-email,143.000)  
    (ft-ftps,104.000)  
    (ft-scp,0.000)  
    (ft-sftp,17.000)
    (ft-skype,187.000) 
    (p2p-torr,233.000) 
    (v-fb,0.000)    
    (v-hng,0.000)   
    (v-ntflx,121.000)  
    (v-skype,0.000) 
    (v-vimeo,96.000)
    (v-ytube,116.000)  
};

\addplot+[
    blue!90!black,fill=blue!35!white,
] coordinates {
    (a-fb,1196.000) 
    (a-hng,7854.000)
    (a-skype,753.000)  
    (a-sptfy,68.000)
    (a-vpbust,1578.000)
    (c-aim,19.000)  
    (c-fb,1100.000) 
    (c-gmail,0.000) 
    (c-hng,2654.000)
    (c-icq,23.000)  
    (c-skype,35.000)
    (e-email,155.000)  
    (ft-ftps,21.000)
    (ft-scp,0.000)  
    (ft-sftp,11.000)
    (ft-skype,668.000) 
    (p2p-torr,242.000) 
    (v-fb,0.000)    
    (v-hng,0.000)   
    (v-ntflx,52.000)
    (v-skype,0.000) 
    (v-vimeo,40.000)
    (v-ytube,97.000)
};

\addplot+[
    yellow!60!black,fill=yellow!35!white,
] coordinates {
    (a-fb,0.000)    
    (a-hng,0.000)   
    (a-skype,0.000) 
    (a-sptfy,0.000) 
    (a-vpbust,1.000)
    (c-aim,0.000)   
    (c-fb,0.000)    
    (c-gmail,0.000) 
    (c-hng,0.000)   
    (c-icq,0.000)   
    (c-skype,0.000) 
    (e-email,0.000) 
    (ft-ftps,0.000) 
    (ft-scp,0.000)  
    (ft-sftp,0.000) 
    (ft-skype,0.000)
    (p2p-torr,0.000)
    (v-fb,0.000)    
    (v-hng,0.000)   
    (v-ntflx,0.000) 
    (v-skype,0.000) 
    (v-vimeo,0.000) 
    (v-ytube,0.000) 
};

\end{axis}

\end{tikzpicture}

%% file: assets/tikz/mappgraph/mg_avg_packet_size.tikz

\begin{tikzpicture}
\definecolor{cpurple}{RGB}{255,0,255}
 
\begin{axis} [ybar=0.1pt, 
height = 4cm, width = 14cm,
bar width = 4pt,
ymin = 1,
ymode=log,
title=\textcolor{blue}{(\ref{app:fig:mg_avg_pkt_size})} AVG Pkt Size x Session,every axis title/.style={below right,at={(0,1)},draw=gray,fill=black!5,font=\fontsize{6}{6}\selectfont},
log basis y={10},
ytick distance=10,
ymajorgrids=true,
enlarge x limits=0.055,
xticklabel shift={-4pt},
yticklabel shift={-2pt},
symbolic x coords={
facebook, twitch, instagram, skype, spotify, google meet, soundcloud, zoom
},
xtick={
facebook, twitch, instagram, skype, spotify, google meet, soundcloud, zoom
},
x tick label style={
font=\fontsize{8}{1}\selectfont
},
y tick label style={font=\tiny},
ytick style={draw=none},
]

\addplot+[
   blue!90!black,fill=blue!35!white,
    error bars/.cd,
    y dir=both,
    y explicit relative,
    error mark options={
      rotate=90,
      mark size=0.9pt,
      line width=0.5pt
    }
] coordinates {
    (facebook,136.431)        +- (0, 1.560)
    (google meet,614.924)     +- (0, 0.326)
    (instagram,202.429)       +- (0, 1.484)
    (skype,264.669)           +- (0, 0.386)
    (soundcloud,133.892)      +- (0, 1.144)
    (spotify,101.695)         +- (0, 1.248)
    (twitch,491.724)          +- (0, 1.264)
    (zoom,626.845)            +- (0, 0.615)
};
\addplot+[
    blue!90!black,fill=blue!60!white,
    error bars/.cd,
    y dir=both,
    y explicit relative,
    error mark options={
      rotate=90,
      mark size=0.9pt,
      line width=0.5pt
    }
] coordinates {
    (facebook,1233.183)       +- (0, 0.150)
    (google meet,592.795)     +- (0, 0.384)
    (instagram,1216.521)      +- (0, 0.147)
    (skype,298.484)           +- (0, 0.150)
    (soundcloud,1375.736)     +- (0, 0.168)
    (spotify,1409.486)        +- (0, 0.210)
    (twitch,1023.854)         +- (0, 0.612)
    (zoom,558.434)            +- (0, 0.687)
};

\end{axis}

\end{tikzpicture}

%% file: assets/tikz/mappgraph/mg_avg_packets.tikz

\begin{tikzpicture}
\definecolor{cpurple}{RGB}{255,0,255}
 
\begin{axis} [ybar=0.1pt, 
height = 4cm, width = 14cm,
bar width = 4pt,
ymin = 1,
ymax=1100000,
ymode=log,
title=\textcolor{blue}{(\ref{app:fig:mg_avg_pkts})} AVG Packets x Session,every axis title/.style={below right,at={(0,1)},draw=gray,fill=black!5,font=\fontsize{6}{6}\selectfont},
log basis y={10},
ytick distance=10,
ymajorgrids=true,
enlarge x limits=0.055,
xticklabel shift={-4pt},
yticklabel shift={-2pt},
symbolic x coords={
facebook, twitch, instagram, skype, spotify, google meet, soundcloud, zoom
},
xtick={
facebook, twitch, instagram, skype, spotify, google meet, soundcloud, zoom
},
x tick label style={
font=\fontsize{8}{1}\selectfont
},
y tick label style={font=\tiny},
ytick style={draw=none},
]

\addplot+[
    blue!90!black,fill=blue!35!white,
    error bars/.cd,
    y dir=both,
    y explicit relative,
    error mark options={
      rotate=90,
      mark size=0.9pt,
      line width=0.5pt
    }
] coordinates {
    (facebook,178.698)        +- (0, 12.858)
    (google meet,876.021)     +- (0, 25.960)
    (instagram,297.505)       +- (0, 7.437)
    (skype,610.407)           +- (0, 15.712)
    (soundcloud,72.115)       +- (0, 4.588)
    (spotify,115.086)         +- (0, 6.609)
    (twitch,1259.639)         +- (0, 17.980)
    (zoom,1041.285)           +- (0, 16.184)
};
\addplot+[
    blue!90!black,fill=blue!60!white,
    error bars/.cd,
    y dir=both,
    y explicit relative,
    error mark options={
      rotate=90,
      mark size=0.9pt,
      line width=0.5pt
    }
] coordinates {
    (facebook,369.023)        +- (0, 12.565)
    (google meet,722.692)     +- (0, 23.374)
    (instagram,679.045)       +- (0, 7.959)
    (skype,820.621)           +- (0, 16.026)
    (soundcloud,147.879)      +- (0, 5.706)
    (spotify,258.797)         +- (0, 6.655)
    (twitch,1292.786)         +- (0, 17.607)
    (zoom,948.065)            +- (0, 15.471)
};

\end{axis}

\end{tikzpicture}

%% file: assets/tikz/mappgraph/mg_avg_bytes.tikz

\begin{tikzpicture}
\definecolor{cpurple}{RGB}{255,0,255}
 
\begin{axis} [ybar=0.1pt, 
height = 4cm, width = 14cm,
bar width = 4pt,
ymin = 1,
ymax=1100000000,
ymode=log,
title=\textcolor{blue}{(\ref{app:fig:mg_avg_bytes})} AVG Bytes x Session,every axis title/.style={below right,at={(0,1)},draw=gray,fill=black!5,font=\fontsize{5}{6}\selectfont},
log basis y={10},
ytick distance=10,
ymajorgrids=true,
enlarge x limits=0.055,
xticklabel shift={-4pt},
yticklabel shift={-2pt},
symbolic x coords={
facebook, twitch, instagram, skype, spotify, google meet, soundcloud, zoom
},
xtick={
facebook, twitch, instagram, skype, spotify, google meet, soundcloud, zoom
},
x tick label style={
font=\fontsize{8}{1}\selectfont
},
y tick label style={font=\tiny},
ytick style={draw=none},
]

\addplot+[
   blue!90!black,fill=blue!35!white,
    error bars/.cd,
    y dir=both,
    y explicit relative,
    error mark options={
      rotate=90,
      mark size=0.9pt,
      line width=0.5pt
    }
] coordinates {
    (facebook,24380.462)      +- (0, 16.418)
    (google meet,538697.175)  +- (0, 30.265)
    (instagram,60231.244)     +- (0, 7.222)
    (skype,161646.358)        +- (0, 16.518)
    (soundcloud,9655.570)     +- (0, 5.812)
    (spotify,11703.931)       +- (0, 7.237)
    (twitch,619401.116)       +- (0, 34.165)
    (zoom,652777.307)         +- (0, 21.949)
};
\addplot+[
    blue!90!black,fill=blue!60!white,
    error bars/.cd,
    y dir=both,
    y explicit relative,
    error mark options={
      rotate=90,
      mark size=0.9pt,
      line width=0.5pt
    }
] coordinates {
    (facebook,455075.393)     +- (0, 12.923)
    (google meet,428408.378)  +- (0, 22.839)
    (instagram,826075.988)    +- (0, 8.154)
    (skype,244945.333)        +- (0, 16.301)
    (soundcloud,203442.058)   +- (0, 6.069)
    (spotify,364771.227)      +- (0, 6.993)
    (twitch,1323626.296)      +- (0, 19.845)
    (zoom,529431.603)         +- (0, 20.547)
};

\end{axis}

\end{tikzpicture}

%% file: assets/tikz/mappgraph/mg_avg_duration_s.tikz

\begin{tikzpicture}
\definecolor{cpurple}{RGB}{255,0,255}
 
\begin{axis} [ybar=0.1pt, 
height = 4cm, width = 14cm,
bar width = 4pt,
ymin = 1,
ymax=11000,
ymode=log,
title=\textcolor{blue}{(\ref{app:fig:mg_avg_dur_s})} AVG Duration (sec) x Session,every axis title/.style={below right,at={(0,1)},draw=gray,fill=black!5,font=\fontsize{6}{6}\selectfont},
log basis y={10},
ytick distance=10,
ymajorgrids=true,
enlarge x limits=0.055,
xticklabel shift={-4pt},
yticklabel shift={-2pt},
symbolic x coords={
facebook, twitch, instagram, skype, spotify, google meet, soundcloud, zoom
},
xtick={
facebook, twitch, instagram, skype, spotify, google meet, soundcloud, zoom
},
x tick label style={
font=\fontsize{8}{1}\selectfont
},
y tick label style={font=\tiny},
ytick style={draw=none},
]

\addplot+[
   blue!90!black,fill=blue!35!white,
    error bars/.cd,
    y dir=both,
    y explicit relative,
    error mark options={
      rotate=90,
      mark size=0.9pt,
      line width=0.5pt
    }
] coordinates {
    (facebook,123.877)        +- (0, 5.779)
    (google meet,129.724)     +- (0, 4.165)
    (instagram,122.818)       +- (0, 3.566)
    (skype,108.047)           +- (0, 4.377)
    (soundcloud,235.356)      +- (0, 4.873)
    (spotify,260.308)         +- (0, 4.515)
    (twitch,185.935)          +- (0, 3.518)
    (zoom,220.345)            +- (0, 4.443)
};
\addplot+[
    blue!90!black,fill=blue!60!white,
    error bars/.cd,
    y dir=both,
    y explicit relative,
    error mark options={
      rotate=90,
      mark size=0.9pt,
      line width=0.5pt
    }
] coordinates {
    (facebook,82.320)         +- (0, 5.601)
    (google meet,113.423)     +- (0, 4.388)
    (instagram,87.668)        +- (0, 3.734)
    (skype,78.227)            +- (0, 4.812)
    (soundcloud,152.612)      +- (0, 4.455)
    (spotify,147.532)         +- (0, 4.931)
    (twitch,160.684)          +- (0, 3.701)
    (zoom,125.670)            +- (0, 5.522)
};

\end{axis}

\end{tikzpicture}

%% file: assets/tikz/mappgraph/mg_avg_iat_ms.tikz

\begin{tikzpicture}
\definecolor{cpurple}{RGB}{255,0,255}
 
\begin{axis} [ybar=0.1pt, 
height = 4cm, width = 14cm,
bar width = 4pt,
ymin = 1,
ymode=log,
title=\textcolor{blue}{(\ref{app:fig:mg_avg_iat_ms})} AVG IAT (ms) x Session,every axis title/.style={below right,at={(0,1)},draw=gray,fill=black!5,font=\fontsize{6}{6}\selectfont},
log basis y={10},
ytick distance=10,
ymajorgrids=true,
enlarge x limits=0.055,
xticklabel shift={-4pt},
yticklabel shift={-2pt},
symbolic x coords={
facebook, twitch, instagram, skype, spotify, google meet, soundcloud, zoom
},
xtick={
facebook, twitch, instagram, skype, spotify, google meet, soundcloud, zoom
},
x tick label style={
font=\fontsize{8}{1}\selectfont
},
y tick label style={font=\tiny},
ytick style={draw=none},
]

\addplot+[
   blue!90!black,fill=blue!35!white,
    error bars/.cd,
    y dir=both,
    y explicit relative,
    error mark options={
      rotate=90,
      mark size=0.9pt,
      line width=0.5pt
    }
] coordinates {
    (facebook,776.235)        +- (0, 35.431)
    (google meet,157.890)     +- (0, 32.945)
    (instagram,436.681)       +- (0, 19.905)
    (skype,184.212)           +- (0, 16.280)
    (soundcloud,3474.621)     +- (0, 12.884)
    (spotify,2386.107)        +- (0, 11.600)
    (twitch,155.938)          +- (0, 33.835)
    (zoom,223.042)            +- (0, 33.710)
};
\addplot+[
    blue!90!black,fill=blue!60!white,
    error bars/.cd,
    y dir=both,
    y explicit relative,
    error mark options={
      rotate=90,
      mark size=0.9pt,
      line width=0.5pt
    }
] coordinates {
    (facebook,253.823)        +- (0, 64.320)
    (google meet,168.082)     +- (0, 32.805)
    (instagram,134.366)       +- (0, 25.741)
    (skype,99.731)            +- (0, 24.718)
    (soundcloud,1105.083)     +- (0, 21.566)
    (spotify,612.079)         +- (0, 24.760)
    (twitch,130.093)          +- (0, 27.806)
    (zoom,137.505)            +- (0, 20.837)
};

\end{axis}

\end{tikzpicture}

%% file: assets/tikz/mappgraph/mg_protocol_dist.tikz
\begin{tikzpicture}

\begin{axis} [ybar=0.1pt,
height = 3.5cm, width = 14cm,
bar width = 4pt,
title=\textcolor{blue}{(\ref{app:fig:mg_proto})} Protocol x Session,every axis title/.style={below left,at={(1,1)},draw=gray,fill=black!5,font=\fontsize{5.5}{6}\selectfont},
label style={font=\small},
ymin = 1,
ymax=110000,
ymode=log,
log basis y={10},
ytick distance=10,
ymajorgrids=true,
enlarge x limits=0.055,
xticklabel shift={-4pt},
yticklabel shift={-2pt},
symbolic x coords={
facebook, twitch, instagram, skype, spotify, google meet, soundcloud, zoom
},
xtick={
facebook, twitch, instagram, skype, spotify, google meet, soundcloud, zoom
},
x tick label style={
font=\fontsize{8}{1}\selectfont
},
y tick label style={font=\tiny},
xtick style={draw=none},
ytick style={draw=none},
]

\addplot+[
    gray!90!black,fill=gray!60!white,
] coordinates {
    (facebook,20)
    (google meet,18)
    (instagram,57)
    (skype,3)
    (soundcloud,10)
    (spotify,19)
    (twitch,19)
    (zoom,19) 
};
\addplot+[
    red!50!gray,fill=red!35!white,
] coordinates {
    (facebook,6)
    (instagram,12)
    (skype,2)
    (soundcloud,8)
    (spotify,9)
    (twitch,6)
    (zoom,21)
};

\addplot+[
    green!90!black,fill=green!35!white,
] coordinates {
    (facebook,7486)
    (google meet,1794)
    (instagram,4307)
    (skype,576)
    (soundcloud,4283)
    (spotify,2759)
    (twitch,5104)
    (zoom,1347)
};

\addplot+[
    blue!90!black,fill=blue!35!white,
] coordinates {
    (facebook,10357)
    (google meet,2749)
    (instagram,6440)
    (skype,712)
    (soundcloud,3128)
    (spotify,3992)
    (twitch,3045)
    (zoom,4597)
};

\addplot+[
    yellow!60!black,fill=yellow!35!white,
] coordinates {
    (facebook,47)
    (google meet,9)
    (instagram,37)
    (skype,10)
    (soundcloud,16)
    (spotify,20)
    (twitch,31)
    (zoom,51)
};

\end{axis}

\end{tikzpicture}

%% file: assets/tikz/mappgraph/mg_unopen_tcp_dist.tikz
\begin{tikzpicture}
 
\begin{axis} [ybar=0.3pt,
height = 4cm, width = 14cm,
bar width = 6pt,
title=\textcolor{blue}{(\ref{app:fig:mg_unopen})} Unopened TCP Sessions,every axis title/.style={below right,at={(0,1)},draw=gray,fill=black!5,font=\fontsize{6}{6}\selectfont},
label style={font=\small},
ymin = 1,
ymode=log,
log basis y={10},
ytick distance=10,
ymajorgrids=true,
enlarge x limits=0.055,
xticklabel shift={-4pt},
yticklabel shift={-2pt},
symbolic x coords={
facebook, twitch, instagram, skype, spotify, google meet, soundcloud, zoom
},
xtick={
facebook, twitch, instagram, skype, spotify, google meet, soundcloud, zoom
},
x tick label style={
font=\fontsize{8}{1}\selectfont,
},
y tick label style={font=\tiny},
xtick style={draw=none},
ytick style={draw=none},
]

\addplot+[
    green!60!black,fill=green!35!white,
] coordinates {
    (facebook,59)
    (google meet,10)
    (instagram,125)
    (skype,159)
    (soundcloud,1)
    (spotify,12)
    (twitch,40)
    (zoom,141)
};

\end{axis}

\end{tikzpicture}